\documentclass{article}
\usepackage{ijcai22}

\usepackage{times}
\usepackage{soul}
\usepackage{url}
\usepackage[hidelinks]{hyperref}
\usepackage[utf8]{inputenc}
\usepackage[small]{caption}
\usepackage{graphicx}
\usepackage{amsmath}
\usepackage{amsthm}
\usepackage{booktabs}
\usepackage{algorithm}
\usepackage{algorithmic}
\usepackage{comment}
\usepackage{amsfonts}
\usepackage{multirow}

\iffalse
\usepackage{color}
\newcommand{\baihua}[1]{{\color{blue}{\bf{Baihua says:}} \emph{#1}}}
\newcommand{\qize}[1]{{\color{red} {#1}}}
\else
\newcommand{\baihua}[1]{}
\newcommand{\qize}[1]{}
\fi

\newtheorem{definition}{Definition}
\newtheorem{problem}{Problem}

\title{Multi-Agent Reinforcement Learning for Traffic Signal Control through Universal Communication Method}

\author{
Qize Jiang$^{1,2,3}$\and
Minhao Qin$^{1,2,3}$\and
Shengmin Shi$^{1,2,3}$\and
Weiwei Sun$^{1,2,3}$\And
Baihua Zheng$^4$
\affiliations
$^1$School of Computer Science, Fudan University\\
$^2$Shanghai Key Laboratory of Data Science, Fudan University\\
$^3$Shanghai Institute of Intelligent Electronics $\&$ Systems\\
$^4$School of Computing and Information Systems, Singapore Management University
\emails
\{qzjiang21,mhqin21\}@m.fudan.edu.cn,
\{cmshi20,wwsun\}@fudan.edu.cn,
bhzheng@smu.edu.sg
}

\begin{document}

\maketitle

\begin{abstract}
%    In recent years, many approaches have been proposed to control traffic signals adaptively.
    %in order to improve traffic congestion and achieved better results
    %compared with conventional methods. 
    How to coordinate the communication among intersections effectively in real complex traffic scenarios with multi-intersection is challenging.
    Existing approaches only enable the communication in a heuristic manner
    %, such as communicating hidden states or observations,
    without considering the content/importance of information to be shared. %communicated. 
    In this paper, we propose a universal communication form \emph{UniComm} between intersections. 
    %With a strong support from theoretical analysis, 
    UniComm embeds massive observations collected at one agent into crucial predictions of their impact on its neighbors, which improves the communication efficiency and is universal across existing methods. 
    %We prove that UniComm contains \qize{lesser} information but can represent the full true state for an agent. 
    %To address the challenge, we propose a cooperative multi-agent reinforcement learning model \emph{UniLight}, which contains a theoretically supported(expand it) communication method, and a corresponding model to make decisions for agents. 
    %On top of UniComm, 
    We also propose a concise network \emph{UniLight} to make full use of communications enabled by UniComm.
    Experimental results on real datasets demonstrate that UniComm universally improves the performance of existing state-of-the-art methods, and UniLight significantly outperforms existing methods on a wide range of
    %both simple and complicated 
    traffic situations.
    Source codes are available at \url{https://github.com/zyr17/UniLight}.
\end{abstract}

\section{Introduction}

%With the development of urbanization and the increasing ownership of cars, 
Traffic congestion is a major problem in modern cities. 
While the use of traffic signal mitigates the congestion to a certain extent, most traffic signals are controlled by timers. 
Timer-based systems are simple, but their performance might deteriorate at intersections with inconsistent traffic volume. 
Thus, an adaptive traffic signal control method, especially controlling multiple intersections simultaneously, is required.
%the key to improve the traffic condition.
%Expanding roads is a common way to mitigate the congestion, 

Existing conventional methods, such as
%such as Self-Organizing Traffic Light Control (SOTL) 
SOTL~\cite{Cools2013}, 
use observations of intersections to form better strategies, but they do not consider long-term effects of different signals and lack a proper coordination among intersections. 
%\baihua{Will audience asks what is agent? We mention agent in Abstract too before formal explanation on agent.}\qize{I've changed `agents' in abstract and here to `intersections'. }
%With the development of artificial intelligence,
Recently, reinforcement learning (RL) methods have showed promising performance in controlling traffic signals. Some of them achieve good performance in controlling traffic signals at a single intersection~\cite{zheng2019learning,DBLP:conf/nips/OroojlooyNHS20}; others focus on collaboration of multi-intersections~\cite{wei2019colight,chen2020toward}.%,  as they can learn a policy to take next actions by observed states directly with reinforcement learning. 

%Besides the achievements reached by exiting RL methods, there are
Multi-intersection traffic signal control is a typical multi-agent reinforcement learning problem. The main challenges include stability, nonstationarity, and curse of dimensionality. 
Independent Q-learning splits the state-action value function into independent tasks performed by individual agents to solve the curse of dimensionality. 
%\baihua{'into independent agents' means decompose the function into independent tasks performed by different agents?} \qize{yes.}
However, given a dynamic environment that is common as agents may change their policies simultaneously, the learning process could be unstable and divergent. 
%
%To increase the stability, 
%Parameter sharing \cite{peng2017multiagent} %provides a solution, as it 
%enables agents to share the same policy, which makes the environment more stable and allows the network to converge faster.
%the convergence speed of the network becomes quicker. 
%

To enable the sharing of information among agents, a proper communication mechanism is required.
%needs to be employed. 
This is critical as it determines the content/amount of the information each agent can observe and learn from its neighboring agents, which directly impacts the amount of uncertainty that can be reduced. 
%The communication mechanism, i.e. some agents share some information with other agents, is also commonly used. 
%With communication between other agents, they can observe more about the true state, and decrease the uncertainty, which leads to better stability.
%For multi-intersection traffic signal control, usually two neighboring agents exchange their information.
Common approaches include enabling the neighboring agents i) to exchange their information with each other and use the partial observations directly during the learning~\cite{el2013multiagent}, or ii) to share hidden states as the information~\cite{wei2019colight,yu2020macar}. 
%Some approaches use the partial observations as the information directly \cite{el2013multiagent,arel2010reinforcement}. %others use hidden states as the information \cite{wei2019colight,yu2020macar}. 
%
While enabling communication is important to stabilize the training process, existing methods have not yet examined the impact of the content/amount of the shared information. 
%to be shared. 
%which data to choose and how to arrange data.
For example, when each agent shares more information with other agents, 
%With bigger data transferred to other agents, %
the network needs to manage a larger number of parameters and hence converges in a slower speed, which actually reduces the stability. 
%the parameter number of network becomes larger, and hence the network converges in a slower speed, which worsens the stability. 
%Consider directly using all neighboring observations of agent $i$, the traffic movements on neighboring intersection $j$ which never passes intersection $i$ makes almost no effect to $i$, but enlarges the input dimension. 
As reported in \cite{zheng2019diagnosing}, additional information does not always lead to better results. 
Consequently, it is very important to select the right information for sharing. 
%The problem also exists for hidden state transition based methods, for the hidden state size is empirically set, and there's no exact meaning for the \qize{elements} in it. 

%While considering the fact that 
Motivated by the deficiency in the current communication mechanism adopted by existing methods, we propose a universal communication form \emph{UniComm}. To facilitate the understanding of UniComm, let's consider two neighboring intersections $I_i$ and $I_j$ connected by a unidirectional road $R_{i,j}$ from $I_i$ to $I_j$.
%Obviously, only $I_i$ could affect $I_j$ and vehicles from $I_i$ to $I_j$ 
If vehicles in $I_i$ impact $I_j$, they
have to first pass through $I_i$ and then follow $R_{i,j}$ to reach $I_j$. 
This observation inspires us to propose UniComm, which picks relevant observations from an agent $A_i$ who manages $I_i$, predicts their impacts on road $R_{i,j}$, and only shares the prediction with its neighboring agent $A_j$ that manages $I_j$. We conduct a theoretical analysis to confirm that UniComm does pass the most important information to each intersection.  
%\baihua{We use information, data and messages in different places. Better to standardize them.}
%
%the fact that when intersection $I_i$ is effecting a neighboring intersection $I_j$, the only way is to effect the unidirectional road $R_{i,j}$ connecting two intersections.
%, which contains the vehicle volume as the observation. 
%Furthermore, the only way to approach the road $R_{i,j}$ is pass though the intersection $I_i$, which is controlled by agent $A_i$. 
%We pick relevant observations from agent $A_i$ and predict its influence to road $R_{i,j}$ as UniComm. Theoretical analysis reveals that UniComm contains the most important messages for intersection $I_j$.

While UniComm addresses the inefficiency of the current communication mechanism, its strength might not be fully achieved by existing methods, whose network structures are designed independent of UniComm. 
%In order to fully utilize the strength of UniComm, 
We therefore design \emph{UniLight}, a concise network structure based on the observations made by an intersection and the information shared by UniComm.
%
%the network structure of existing methods is not good enough to make full use of it. 
%So we design a concise network based on the observation and UniComm, which is named as \emph{UniLight}. 
%UniLight 
It predicts the Q-value function of every action based on the importance of different traffic movements.
%UniLight achieves best performance in public datasets, as well as two newly generated datasets, which are more complicated. 

In brief, we make three main contributions in this paper. %\textbf{Firstly}, we propose a universal communication form \emph{UniComm} in multi-intersection traffic signal control problem, with its effectiveness supported by a thorough  theoretical analysis. \textbf{Secondly}, we propose a traffic movement importance based network \emph{UniLight} to make full use of observations and UniComm. \textbf{Last but not least}, we conduct experiments to demonstrate that UniComm is universal for all existing methods, and UniLight can achieve superior performance on both simple and complicated datasets. 
%
%\begin{itemize}
%    \item 
Firstly, we propose a universal communication form \emph{UniComm} in multi-intersection traffic signal control problem, with its effectiveness supported by a thorough  theoretical analysis.
%    \item 
Secondly, we propose a traffic movement importance based network \emph{UniLight} to make full use of observations and UniComm.
%    \item 
Thirdly, we conduct experiments to demonstrate that UniComm is universal for all existing methods, and UniLight can achieve superior performance on not only simple but also complicated traffic situations.
%\end{itemize}

\section{Related Works}

%Conventional adaptive traffic signal control methods mainly consider single intersection controlling based on its observations. SOTL\cite{???} and MaxPressure\cite{???} evaluates the congestion of every traffic phase by queue lengths of the intersection, and choose whether to switch the phase. For multiple intersection controlling methods, such as GreenWave\ref{???}, changes the offset of different intersections to make less stopping count. 

Based on the number of intersections considered, traffic signal control problem (TSC) can be clustered into i) single intersection traffic signal control (S-TSC) and ii) multi-intersection traffic signal control (M-TSC). 
%We also briefly introduce deep reinforcement learning, as they have been used widely to solve TSC problems. 

%\paragraph{Single-intersection traffic signal control.}
%\paragraph{S-TSC.}
\noindent\textbf{S-TSC.}
%Controlling traffic signal at one intersection 
S-TSC is a sub-problem of M-TSC, as decentralized multi-agent TSC is widely used. 
Conventional %adaptive controlling 
methods like SOTL %(e.g., SOTL \cite{Cools2013}) 
choose the next phase by current vehicle volumes with limited flexibility.  
%but such a heuristic method lacks flexibility to adapt to different situations and conveniently ignores the cumulative impact of the past observations on the selections of current phases. %\baihua{Cumulative impact of which kind of information on what?}
%
As TSC could be modelled as a Markov Decision Process (MDP), many recent methods adopt reinforcement learning (RL)~\cite{DBLP:conf/aaai/HasseltGS16,DBLP:conf/icml/MnihBMGLHSK16,DBLP:conf/icml/HaarnojaZAL18,ault2020learning}.
In RL, agents interact with the environment and take rewards from the environment, and different algorithms are proposed to learn a policy that maximizes the expected cumulative reward received from the environment.
Many algorithms~\cite{zheng2019learning,zang2020metalight,DBLP:conf/nips/OroojlooyNHS20}
%Reinforcement learning (RL) based methods have become popular recently, because of their ability to learn and be adaptive, e.g.,  \cite{zheng2019learning,zang2020metalight,DBLP:conf/nips/OroojlooyNHS20}.
%\cite{li2016traffic} uses stacked auto-encoders to approximate rewards; \cite{wei2018intellilight} controls two-phase traffic signal with deep learning; \cite{zheng2019learning} proposes FRAP method to fit intersections having different shapes; \cite{zhang2020using} tries to control traffic signals when only part of vehicles can be detected; \cite{zang2020metalight,oroojlooy2020attendlight} use meta learning and attention methods to improve the training speed and model generalizability. 
%\cite{oroojlooy2020attendlight} uses attention method to perform better results and generalizability. 
%Although they 
though perform well for S-TSC, their performance at 
M-TSC is not stable, as they suffer from a poor generalizability and their models are hard to train. 
%but 
%not perform well with multiple intersections. %
%because they lacks of generalizability or hard to train, they are not stable to achieve a good result with multiple intersections. 
%\qize{because they have blabla problem, can only use in single.} Some of them lacks generalizability to learn strategies that fit different intersections, and some of them are hard to train, which leads to nonstationarity and can't learn stable strategies. 

%\paragraph{Multi-intersection traffic signal control.}
%\paragraph{M-TSC.}
%\paragraph{Conventional multi-intersection traffic signal control.}
\noindent\textbf{M-TSC.}
Conventional methods for M-TSC mainly coordinate different traffic signals by changing their offsets, which 
%has a rather limited impact on efficiency, and 
only works for a few pre-defined directions and has low efficiency.
%Recently, MaxPressure~\cite{varaiya2013max} has tried to minimize the pressure metric in order to equalize every intersection's pressure, which is proved to be equivalent to minimize vehicle's passing time.
%
%\paragraph{RL-based multi-intersection traffic signal control.}
When adapting RL based methods from single intersection to multi-intersections, we can treat every intersection as an independent agent. 
However, due to the unstable and dynamic nature of the environment, 
%But because the environment is unstable and nonstationarity, 
the learning process is hard to converge \cite{bishop2006pattern,nowe2012game}. 
Many methods have been proposed to speedup the convergence, including parameter sharing 
and approaches that design different rewards to contain neighboring information~\cite{chen2020toward}. 
%Considering cooperate by communication, \cite{el2013multiagent,arel2010reinforcement} directly communicate by neighboring observations to improve the stability.
%But this method has the side effect that the final observation space rises significantly.
%
Agents can also communicate with their neighboring agents via either direct or indirect communication. 
The former is simple but results in a very large observation space~\cite{el2013multiagent,arel2010reinforcement}. The latter relies on the learned hidden states and many different methods~\cite{nishi2018traffic,wei2019colight,chen2020toward} have been proposed to facilitate a better generation of hidden states and a more cooperative communication among agents. 
%
%Other methods try to communicate with neighboring agents by learned hidden states. 
%To better generating the hidden state and cooperate between agents, many different methods have been proposed.
%\cite{yu2020macar,nishi2018traffic,wei2019colight} use graph structure based neural networks Message Propagation Graph Neural Network, Graph Attention Network and Graph Convolutional Network; \cite{zhu2021meta} uses meta learning and intrinsic reward to learn generalized model; %\cite{wei2019colight} proposed CoLight, which uses attention mechanism to gather information from neighbors with learnable weight. \cite{chen2020toward} combines FRAP and PressLight. %, and tests on the environment with 2,510 intersections. 
While many methods show good performance in experiments, their communication is mainly based on hidden states extracted from neighboring intersections. They neither examine the content/importance of the information, nor consider what is the key information that has to be passed from an agent to a neighboring agent. 
%what is the real effect between two neighboring intersections.
%\baihua{What do you mean by "the real effect between two neighboring intersections"?}
This makes the learning process of hidden states more difficult, and models may fail to learn a reasonable result when the environment is complicated. 
Our objective is to develop a communication mechanism that has a solid theoretical foundation and meanwhile is able to achieve a good performance.% in empirical evaluations.  

\section{Problem Definition}

We consider M-TSC as a Decentralized Partially Observable Markov Decision Process (Dec-POMDP)~\cite{oliehoek2016concise},
which can be described as a tuple $\mathcal{G} = \langle \mathcal{S}, \mathcal{A}, P, r, \mathcal{Z}, O, N, \gamma \rangle$. Let $\boldsymbol{s}\in\mathcal{S}$ indicate the current true state of the environment. 
Each agent $i\in\mathcal{N}:={1,\cdots,N}$ chooses an action $a_i\in \mathcal{A}$, with  $\boldsymbol{a}:=[a_i]_{i=1}^N\in\mathcal{A}^N$ referring to the joint action vector formed.
The joint action then transits the current state $\boldsymbol{s}$ to another state $\boldsymbol{s}'$, according to the state transition function $P(\boldsymbol{s}'|\boldsymbol{s}, \boldsymbol{a}):\mathcal{S}\times\mathcal{A}^N\times\mathcal{S}\to[0,1]$. 
The environment gets the joint reward by reward function $r(\boldsymbol{s}, \boldsymbol{a}):S\times\mathcal{A}^N\to\mathbb{R}$. %, with the discount factor $\gamma$.  
Each agent $i$ can only get partial observation $z\in\mathcal{Z}$ according to the observation function $O(\boldsymbol{s}, i):\mathcal{S}\times i\to\mathcal{Z}$. 
The objective of all agents is to maximize the cumulative joint reward $\sum_{i=0}^{\infty} \gamma^i r(\boldsymbol{s}_i, \boldsymbol{a}_i)$, where $\gamma\in[0,1]$ is the discount factor.
%Each agent also has an action-observation history $\tau^a\in\Tau:=$

\begin{figure}
    \centering
    \includegraphics[width=0.483\textwidth]{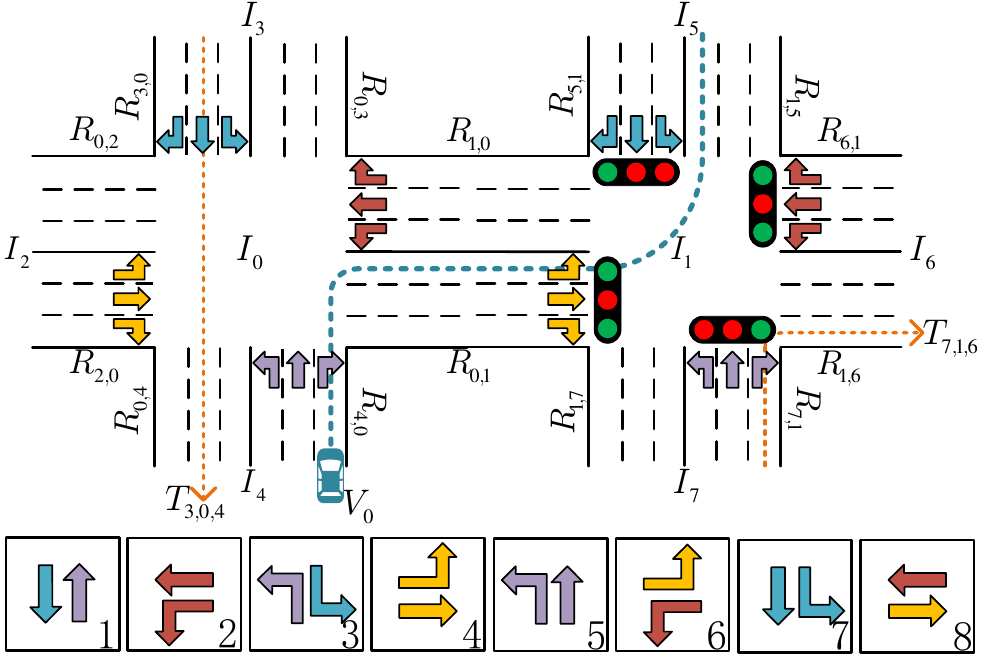}
    %\vspace{-0.1in}
    \caption{Visualization of two adjacent 4-arm intersections and their corresponding definitions, and 8 phases. Phase \#6 is activated in $I_1$. We omit the turn-right traffic movements in all phases as they are always permitted in countries following right-handed driving.
	%\baihua{Make the left sub-figure smaller, so the entire figure will occupy less space.}\qize{it's a sketch now, and I will make it pretty later.}
	}
    \label{fig:definition}
    %\vspace{-5mm}
\end{figure}

Following CoLight~\cite{wei2019colight} and MPLight~\cite{chen2020toward}, we define M-TSC in Problem 1. 
We plot the schematic of two adjacent 4-arm intersections in Figure~\ref{fig:definition} 
to facilitate the understanding of following definitions. 
%more intuitively.
%
%We visualize two adjacent 4-arm intersections in Figure \ref{fig:definition} to explain definitions clearer. 

\begin{definition}
    An \emph{intersection} $I_i\in\mathcal{I}$ refers to the start or the end of a road. 
    If an intersection has more than two approaching roads, %\baihua{Is the number of approaching road the key or the direction of approach lanes the key? If an intersection has two approaching lanes with the same direction - go straight, does it require a traffic light?}\qize{As intersection with two approaching roads can simply consider as middle point of one single road (without U-turns), Is it necessary to mention? I've added description about it, but I think it's ugly...}, 
    it is a real intersection $I^R_i\in\mathcal{I}^R$ as it has a traffic signal. We assume that no intersection has exactly two approaching roads, as both approaching roads have only one outgoing direction and the intersection could be removed by connecting two roads into one.
    %, and we can simply remove the intersection and connect the corresponding roads
    %two approaching roads with two exiting roads as two roads. 
    %\baihua{double check the above statement. Do you mean two approaching roads, or one approaching road and one existing road???}
    If the intersection has exactly one approaching road, it is a virtual intersection $I^V_i\in\mathcal{I}^V$, which usually refers to the border intersections of the environment, such as $I_2$ to $I_7$ in Figure \ref{fig:definition}. %\qize{If has space, maybe add one figure here to show definitions clearly? I have one in my master thesis, but the space may not enough.}
    %For every $I^R_i$, an agent $i$ controls the traffic signal on it. 
    The neighboring intersections $\mathcal{I}^N_i$ of $I_i$ is defined as $\mathcal{I}^N_i = \{I_j|R_{i,j}\in\mathcal{R}\}\cup \{I_j|R_{j,i}\in\mathcal{R}\}$, where roads $R_{i,j}$ and $\mathcal{R}$ are defined in Definition \ref{def:road}. %\baihua{The neighboring intersections refer to the next intersections but not previous intersections. Do we need to mention why we only consider next intersections only?}\qize{I fixed the equation by adding previous intersections. However, although I defined the road as unidirectional for convenience using in proofs, I assumed $R_{i,j}$ must have $R_{j,i}$. I'll check later if there is any conflict in proofs. }
\end{definition}

\begin{definition}
    \label{def:road}
    A \emph{Road} $R_{i,j}\in \mathcal{R}$ is a unidirectional edge from intersection $I_i$ to another intersection $I_j$. 
    $\mathcal{R}$ is the set of all valid roads.
    We assume each road has multiple lanes,
    %There are several lanes on the road 
    and each lane belongs to exactly one traffic movement, which is defined in Definition \ref{def:movement}.
\end{definition}

\begin{definition}
    \label{def:movement}
    A \emph{traffic movement} $T_{x, i, y}$ is defined as the traffic movement travelling across $I_i$ from entering lanes on road $R_{x,i}$ to exiting lanes on road $R_{i,y}$. %\baihua{Should it be from the exiting lanes on $I_i$ to entering lanes on $R_{i,y}$?}\qize{The entering lanes and exiting lanes is considered for $I_i$. lanes approaching $I_i$ is entering lanes, leaving $I_i$ is exiting lane. i.e. lanes on $R_{i,y}$ is exiting lanes for $I_i$, as well as entering lanes for $I_j$. }
    For a 4-arm intersection, there are 12 traffic movements. %(e.g., $T_{a, i, c}$ and $T_{b, i, c}$).
    We define the set of traffic movements passing $I_i$ as $\mathcal{T}_i = \{T_{x,i,y}|x, y \in \mathcal{I}^N_i, R_{x,i}, R_{i,y}\in\mathcal{R}
    %, R_{i,y}\in\mathcal{R}
    \}$.
    $T_{3,0,4}$ and $T_{7,1,6}$ represented by orange dashed lines in Figure \ref{fig:definition} are two example traffic movements.
    %$T_{3,0,4}$ and $T_{7,1,6}$ are shown in orange dashed arrow lines. 
    %\baihua{Double check: $x, y \in \mathcal{I}^N_i$. Based on my understanding, $y \in \mathcal{I}^N_i$ while $i\in \mathcal{I}^N_x$???}
\end{definition}

%Here we omit the rarely appeared U-turn traffic movement for convenience. 

\begin{definition}
    A \emph{vehicle route} is defined as a sequence of roads $V$ with a start time $e\in\mathbb{R}$ that refers to the time when the vehicle enters the environment. 
    Road sequence $V=\langle R^1, R^2, \cdots, R^n\rangle$, with a traffic movement $T$ from $R^i$ to $R^{i+1}$ for every $i\in[1,n-1]$. 
    We assume that all vehicle routes start from/end in virtual intersections. 
    %Start time $e$ refers to the time when the vehicle enters the environment. 
    $\mathcal{V}$ is the set of all valid vehicle routes.
    $V_0$ in Figure \ref{fig:definition} is one example. 
\end{definition}

\begin{definition}
    A traffic signal phase $P_i$ is defined as a set of permissible traffic movements at $I_i$. 
    %\baihua{Do we need to highlight that a phase refers to the set of traffic movements that are permissible at the same time at $I_i$?}
    The bottom of Figure \ref{fig:definition} shows eight phases.
    $\mathcal{A}_i$ denotes the complete set of phases at $I_i$, i.e., the action space for the agent of $I_i$. 
\end{definition}

\begin{problem}
    In multi-intersection traffic signal control (\emph{M-TSC}), the environment consists of intersections $\mathcal{I}$, roads $\mathcal{R}$, and vehicle routes $\mathcal{V}$. 
    Each real intersection $I^R_i\in \mathcal{I}^R$ is controlled by an agent $A_i$. Agents perform actions between time interval $\Delta t$ based on their policies $\pi_i$. 
    At time step $t$, $A_i$ views part of the environment $z_i$ as its observation, and tries to take an optimal action $a_i\in \mathcal{A}_i$ (i.e., a phase to set next) that can maximize the cumulative joint reward $r$.
    %such that the cumulative joint reward $r$ could be maximized.
\end{problem}
%traffic situation and current traffic signal phase
As we define the M-TSC problem as a Dec-POMDP problem, we have the following RL environment settings.

%\begin{itemize}
%    \item 
\noindent\textbf{True state $\boldsymbol{s}$ and partial observation $\boldsymbol{z}$.} 
    At time step $t \in \mathbb{N}$, agent $A_i$ has the partial observation $z_i^t \subseteq \boldsymbol{s}^t$, which contains the average number of vehicles $n_{x,i,y}$ 
    %and average queue length $q_j$ 
    following traffic movement $T_{x,i,y}\in\mathcal{T}_i$ and the current phase $P_i^t$.
    %The true state $\boldsymbol{s_t}$ in time $t$ is the combination of traffic movement information and traffic signal phase information. 
    %The former contains the number of vehicles $\boldsymbol{v_t}$ and the number of waiting vehicles $\boldsymbol{w_t}$ on every lanes, 
    %here $\boldsymbol{w_t}$ means the vehicles waiting in the queue. 
    %which means the vehicle queue length waiting for the traffic signal.
    %The phase information contains current phase $\boldsymbol{p_t}$ selected in every intersections. 
    %For partial observation $z_i$ for agent $i$, there are the number of vehicles and waiting vehicles on road lanes connected to the intersection, and the phase selected by the intersection. To perform better cooperation, every agent will also receive some information generated by neighboring agents.

%    \item 
\noindent\textbf{Action $\boldsymbol{a}$.} After receiving the partial observation $z_i^t$, agent $A_i$ chooses an action $a_i^t$ from its candidate action set $\mathcal{A}_i$ corresponding to a phase in next $\Delta t$ time. 
    %Consistent with the reality, when 
    If the activated phase $P_i^{t+1}$ is different from current phase $P_i^t$, a short all-red phase will be added to avoid collision.

%    \item 
\noindent\textbf{Joint reward $\boldsymbol{r}$.} In M-TSC, we want to minimize the average travel time, which is hard to optimize directly, so some alternative metrics are chosen as immediate rewards. 
    In this paper, we set the joint reward $r^t=\sum_{I_i\in \mathcal{I}^R} r_i^t$, 
    %In this paper, we set the joint reward $r^t$ as the summation $\sum_{I_i\in \mathcal{I}^R} r_i^t$ of rewards received by each intersection $I_i$, %$\sum r_i^t$
    where 
    %and the reward received for intersection $I_i$ is 
    $r_i^t = -\overline{n^{t+1}_{x,i,y}}, s.t.\ x,y\in\mathcal{I}^N_i \land T_{x, i, y}\in\mathcal{T}_i$ indicates the reward received by $I_i$, with $n_{x, i, y}$ the average vehicle number on the approaching lanes of traffic movement $T_{x, i, y}$. 
    %\baihua{In Eq(1), $r_i^t$ is defined based on $n^{t+1}_{x,i,y}$ instead of $n^t_{x,i,y}$. }
    %\qize{Yes, it should be $t+1$. $r^t$ refers to reward for $a^t$, and it is evaluated by $n^{t+1}$ after taking the action.}
    %i.e. the average vehicle number of all traffic movements in $\mathcal{T}_i$. %, where $n_{x, i, y}$ is the average vehicle number on the approaching lanes of traffic movement $T_{x, i, y}$.
%\end{itemize}

\section{Methods}

% \qize{I will check equations and variables, and may rephrase whole section in Jan 12, this part is currently kep unchanged. }

\begin{figure*}[t]
	\centering
	\includegraphics[width=175mm]{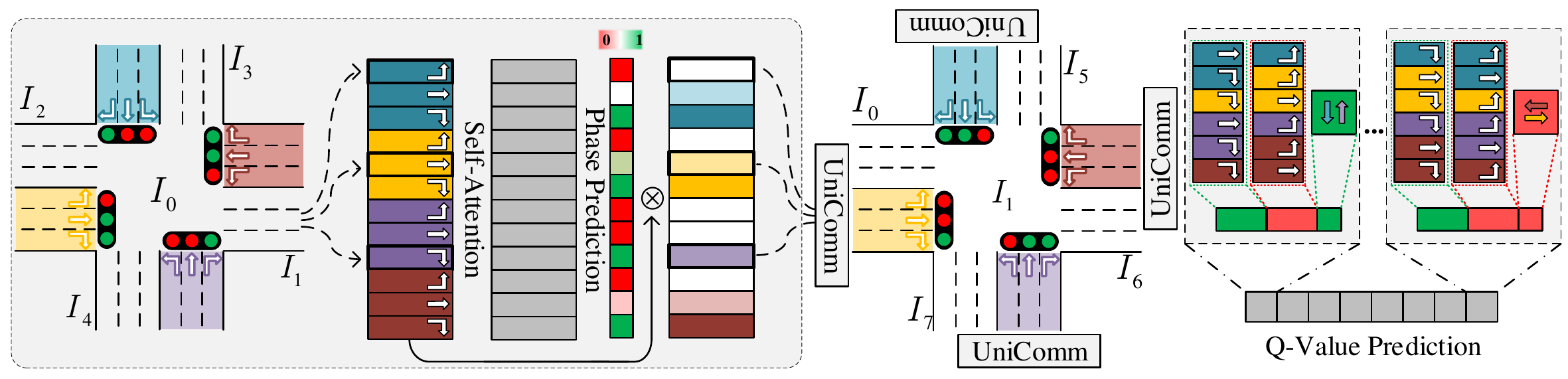}
	%\caption{Available activated traffic flow combinations.}
	%\vspace{-0.05in}
	\caption{UniComm and UniLight structure}
	\label{fig:model}
	%\vspace{-0.2in}
\end{figure*}

In this section, 
%As we mentioned above, 
we first propose UniComm, a new communication method, to improve the communication efficiency among agents; we then construct UniLight, a new controlling algorithm, to control signals with the help of UniComm. 
We use Multi-Agent Deep Q Learning \cite{mnih2015human} with double Q learning and dueling network as the basic reinforcement learning structure.
Figure~\ref{fig:model} illustrates the newly proposed model.%, and the pseudo codes of UniComm and UniLight are presented in the supplementary material.
%The illustration of our proposed model is in Figure~\ref{fig:model}. We also provide the pseudo code of our proposed algorithms in the supplementary material.

%Here we propose our method UniLight. As we mentioned above, there are two critical issues in multi-intersection traffic signal control problem. 
%The first is low convergence speed and overfitting for a single agent, which is caused by not well-designed network structure and large network size. 
%The second problem is the communication efficiency between agents, normal embedding delivery is hard for agents to learn to find important information with only reward as the target. 
%To tackle these two issues, we use two part, A and B, to solve these two corresponding issues.

\subsection{UniComm}\label{subsec:unicomm}

The ultimate goal to enable communication between agents is to mitigate the nonstationarity caused by decentralized multi-agent learning. 
Sharing of more observations could help improve the stationarity, but it meanwhile suffers from the curse of dimensionality. 
%However, when the stationarity is improved, the curse of dimensionality occurs again.
%With more observation received from other agents, the stationarity is improved; however, the curse of dimensionality occurs again.
%For example, if agent $A_i$ can get full state $\boldsymbol{s}$ and policies $\boldsymbol{\pi}$ of other agents,
%, as the policies can be regarded as part of the state, 
%the environment could achieve stable transition probability $P$. However, the observation space could be as large as a single agent to perform joint actions. 
%\baihua{larger than a single agent???}
%\qize{more obesrvation, less stationarity, vise versa?}

%Existing methods use hidden states or observations to communicate between two neighboring agents.
In existing methods, neighboring agents exchange hidden states or observations via communication. 
However, as mentioned above, there is no theoretical analysis on how much information is sufficient and which information is important.
%
%nobody theoretically concerns how much information is enough, and which information is important to communicate.
While with deep learning and back propagation method, we expect the network to be able to recognize information importance via well-designed structures such as attention. 
Unfortunately, as the training process of reinforcement learning is not as stable as that of supervised learning, the less useful or useless information might affect the convergence speed significantly. 
Consequently, how to enable agents to communicate effectively with their neighboring agents becomes critical. 
%the potential impact of useless information to the convergence speed is more significant. 
%So communicating effectively to other agents is important to improve performance. 
%\qize{an example based on pics in introduction}
Consider intersection $I_0$ in Figure \ref{fig:definition}. Traffic movements such as $T_{4, 0, 2}$ and $T_{3, 0, 4}$
will not pass $R_{0, 1}$, and hence their influence to $I_1$ is much smaller than $T_{2, 0, 1}$. 
Accordingly, we expect the information related to $T_{4, 0, 2}$ and $T_{3, 0, 4}$ 
to be less relevant (or even irrelevant) to $I_1$, as compared with information related to $T_{2, 0, 4}$. 
In other words, when $I_0$ communicates with $I_1$ and other neighboring intersections, 
ideally $I_0$ is able to pass different information to different neighbors. 

%To enable effective communication between agents, 
We propose to share only important information with neighboring agents.
%We try to design a communication form which contains only important information. 
To be more specific, we focus on agent $A_1$ on intersection $I_1$ which learns maximizing the cumulative reward of $r_1$ via reinforcement learning and evaluate the importance of certain information based on its impact on $r_1$.
%When context is clear, we omit the subscript ``1'' for simplicity. 
%Consider agent $A_i$, while reinforcement learning learns the cumulative reward of $r_i$, we consider the impact of observations and policies to $r_i$.

We make two assumptions in this work. First, spillback that refers to the situation where a lane is fully occupied by vehicles and hence other vehicles cannot drive in
    never happens. This simplifies our study as spillback
    %This allows us to skip spillback for simplicity, as it 
    rarely happens and will disappear shortly even when it happens. Second, within action time interval $\Delta t$, no vehicle can pass though an entire road.
    This could be easily fulfilled, because $\Delta t$ in M-TSC is usually very short (e.g. $10s$ in our settings).

%\begin{assumption}
%    \label{ass:spillback}
%    Spillback that refers to the situation where a lane is fully occupied by vehicles and hence other vehicles cannot drive in    never happens.
    %, i.e. the situation that a lane is fully occupied by vehicles and prevents other vehicles driving in never happens.
%\end{assumption}
%
%\begin{assumption}
%    \label{ass:passroad}
%    Within action time interval $\Delta t$, no vehicle can pass though an entire road. 
%\end{assumption}

%it is worth to mention that by adding some specified observations, we can also deal with spillback situation. But 

%For assumption \ref{ass:spillback}, as spillback rarely happens, and may disappear after a while, we ignore it here for simplicity. 
%For assumption \ref{ass:passroad}, while $\Delta t$ in traffic signal control problem is usually short (e.g. $10s$ in our settings), the assumption is easily fulfilled. 

Under these assumptions, we decompose reward $r^t_1$ as follows. 
%As $|\boldsymbol{T}_i|$ is constant, for convenience, we use the sum of $n_j$ instead of the average.
Recall that the reward $r^t_1$ is defined as the average of $n_{x,1,y}$ with $T_{x,1,y}\in\mathcal{T}_1$ being a traffic movement. 
As the number of traffic movements $|\mathcal{T}_1|$ w.r.t. intersection $I_1$ is a constant, for convenience, we analyze the sum of $n_{x,1,y}$, i.e. $|\mathcal{T}_1|r_1$, and $r^t_1$ can be %easily obtained 
derived by dividing the sum by $|\mathcal{T}_1|$. 
%Let $C_i=|\mathcal{T}_i|$.
%
\begin{align}
|\mathcal{T}_1|r^t_1 &= \sum{n^{t+1}_{x,1,y}} &x, y \in \mathcal{I}^N_1 \nonumber \\
          &= \sum{\left((n^t_{x,1,y} - m^t_{x,1,y})+ l^t_{x,1,y}\right)} \label{eq:dec} \\
          &= f(z^t_1) + g(\boldsymbol{s}^t) \label{eq:fandg}
\end{align}

In Eq.~(\ref{eq:dec}), we decompose $n^{t+1}_{x, 1, y}$ into three parts, i.e. current vehicle number $n^t_{x, 1, y}$, approaching vehicle number $l^t_{x, 1, y}$, and leaving vehicle number $m^t_{x, 1, y}$. 
Here, $n^t_{x, 1, y} \in z^t_1$ can be observed directly.
All leaving vehicles in $m^t_{x,1,y}$ are on $T_{x,1,y}$ now and can drive into next road without the consideration of spillback (based on first assumption).
$P^t_1$ is derived by $\pi_1$, which uses partial observation $z^t_1$ as input.
Accordingly, $m^t_{x,1,y}$ is only subject to $n^t_{x,1,y}$ and $P^t_1$ that both are captured by the partial observation $z^t_1$.
Therefore, approaching vehicle number $l^t_{x,1,y}$ is the \textbf{only} variable affected by observations $o\notin z^t_1$. We define $f(z^t_1) = \sum{(n^t_{x,1,y} - m^t_{x,1,y})}$ and $g(\boldsymbol{s}^t) = \sum{l^t_{x,1,y}}$, as shown in Eq.~(\ref{eq:fandg}).

To help and guide an agent to perform an action, agent $A_1$ will receive communications from other agents. Let $c^t_1$
denote the communication information received by $A_1$ in time step $t$. In the following, we analyze the impact of $c^t_1$ on $A_1$. 
%We define the communication information received in time step $t$ as $c^t_i$, and consider its impact to $A_i$. 

For observation of next phase $z^{t+1}_1 = \{n^{t+1}_{x,1,y}, P^{t+1}_1\}$, we have i)  $n^{t+1}_{x,1,y} = n^t_{x,1,y} + l^t_{x,1,y} - m^t_{x,1,y}$; and ii) $P^{t+1}_1 = \pi_1(z^t_1, c^t_1)$.
%
%\begin{align*}
%    n^{t+1}_{x,i,y} &= n^t_{x,i,y} + l^t_{x,i,y} - m^t_{x,i,y} \\
%    P^{t+1}_i &= \pi_i(z^t_i, c^t_i)
%\end{align*}
%
As $n^t_{x,1,y}$ and $m^t_{x,1,y}$ are part of $z^t_1$, there is a function $Z_1$ such that $z^{t+1}_1 = Z_1(z^t_1, c^t_1, l^t_{x,1,y})$.
Consequently, to calculate $z^{t+1}_1$, the knowledge of $l^t_{x,1,y}$, that is not in $z^t_1$, becomes necessary. 
Without loss of generality, we assume $l^t_{x,1,y}\subseteq c^t_1$, 
%and we omit $l^t_{x,i,y}$ while $c^t_i$ appears, 
so $z^{t+1}_1 = Z_1\left(z^t_1, c^t_1\right)$. In addition, $z^{t+j}_1$ can be represented as $Z_j(z^t_1, c^t_1, c^{t+1}_1, \cdots, c^{t+j-1}_1)$. 
Specifically, we define $Z_0(\cdots)=z^t_1$, regardless of the input.

We define the cumulative reward of $A_1$ as $R_1$, which can be calculated as follows: 
%\baihua{what is $\gamma^j$?}\qize{It is defined in Dec-POMDP, but I forgot to use it in definition. I've added the definition about cumulative reward. }\baihua{Still could not find it. }\qize{in Line 347, defined cumulative reward, which uses the discount factor $\gamma$. Here $\gamma^j$ is just $\gamma$ to the $j$ power.}
%
\begin{align}
    R^t_1 &= \sum\nolimits_{j=0}^\infty{\gamma^j r^{t+j}_1} = \sum\nolimits_{j=0}^\infty{\gamma^j\left(f\left(z^{t+j}_1\right)+g\left(\boldsymbol{s}^{t+j}\right)\right)} \nonumber \\
%\end{align*}
%\begin{align} 
    &= \sum\nolimits_{j=0}^\infty{\gamma^j\left(f\left(Z_j\left(z^t_1, c^t_1, \cdots, c^{t+j-1}_1\right)\right)+g\left(\boldsymbol{s}^{t+j}\right)\right)} \nonumber %\label{eq:cumulative}
\end{align}

From %Eq.~(\ref{eq:cumulative}),
above equation, we set $c_1^t$ as the future values of $l$:
\begin{align}
\label{eq:unicomm}
c_1^t &= \left\{g\left(\boldsymbol{s}^t\right),g\left(\boldsymbol{s}^{t+1}\right), \cdots \right\} = \left\{\sum{l_{x,1,y}^t}, \sum{l_{x,1,y}^{t+1}}, ... \right\} \nonumber
\end{align}
%
%In addition, we know
All other variables to calculate $R_1^t$ can be derived from $c_1^t$:
\begin{align}
    &c_1^{t+j} = c_1^t\backslash\{g\left(\boldsymbol{s}^t\right), \cdots, g\left(\boldsymbol{s}^{t+j-1}\right)\} & j&\in\mathbb{N}+ \nonumber \\
    &g\left(\boldsymbol{s}^{t+k}\right) \in c_1^t & k&\in\mathbb{N}\nonumber
\end{align}

Hence, it is possible to derive the cumulative rewards $R_1^t$ based on future approaching vehicle numbers of traffic movements $l^{t}_{x, 1, y}$s from $c_1^t$, even if other observations remain unknown. %\baihua{it's not so straightforward to me, why we declare it is possible/practical to calculate the cumulative rewards based on future approaching vehicle numbers of traffic movements.}
As the conclusion is \emph{universal} to all existing methods on the same problem definition regardless of the network structure, we name it
%this communication method 
as \emph{UniComm}. 
Now we convert the problem of finding the cumulative reward $R^t_1$ to how to calculate approaching vehicle numbers $l^t_{x, 1, y}$ with current full observation $\boldsymbol{s}^t$. 
As we are not aware of exact future values, we use existing observations in $\boldsymbol{s}^t$ to 
%perform our prediction to 
predict $l^{t+1}_{x, 1, y}$. 

We first calculate approaching vehicle numbers on next interval $\Delta t$.
%TODO
Take intersections $I_0, I_1$ and road $R_{0,1}$ for example,
for traffic movement $T_{0, 1, x}$ which passes intersections $I_0$, $I_1$, and $I_x$, approaching vehicle number $l_{0, 1, x}$ depends on the traffic phase $P_0^{t+1}$ to be chosen by $A_0$. 
Let's revisit the second assumption. %~\ref{ass:passroad}.
All approaching vehicles should be on $\boldsymbol{T}_{0,1}=\{T_{y,0,1}|T_{y,0,1}\in\mathcal{T}_0, y\in\mathcal{I}^N_0\}$, which belongs to $z^t_0$. 
As a result, $\boldsymbol{T}_{0, 1}$ and the phase $P^{t+1}_0$ affect $l_{0, 1, x}$ the most, even though there might be other factors.
%So although there exists effect from other observations, $\boldsymbol{T}_{i,j}$ and the phase $P^{t+1}_i$ has the most effect to $m_{i,j,k}$. 

We convert observations in $\mathcal{T}_0$ into hidden layers $\boldsymbol{h}_0$ for traffic movements by a fully-connected layer and ReLU activation function. 
As $P^{t+1}_0$ can be decomposed into the permission for every $T\in\mathcal{T}_0$, we use self-attention mechanism~\cite{DBLP:conf/nips/VaswaniSPUJGKP17} between $\boldsymbol{h}_0$ with Sigmoid activation function to predict the permissions $\boldsymbol{g}_0$ of traffic movements in next phase, which directly multiplies to $\boldsymbol{h}_0$. %\baihua{Is $\boldsymbol{g}_i$ 0 or 1?}\qize{it has $|\mathcal{T}_i|$ elements, every element $g_{i,j}\in(0,1)$.}
This is because for traffic movement $T_{x,0,y} \in \mathcal{T}_0$, if $T_{x,0,y}$ is permitted, $h_{x,0,y}$ will affect corresponding approaching vehicle numbers $l_{0,y,z}$; otherwise, it will become an all-zero vector and has no impact on $l_{0,y,z}$.
Note that $R_{x,0}$ and $R_{0,y}$ might have different numbers of lanes, so we scale up the weight by the lane numbers to eliminate the effect of lane numbers.
Finally, we add all corresponding weighted hidden states together and use a fully-connected layer to predict $l_{0, y, z}$. 

To learn the phase prediction $P_0^{t+1}$, a natural method is using the action $\boldsymbol{a}_0$ finally taken by the current network. 
However, as the network is always changing, even the result phase action $\boldsymbol{a}_0$ corresponding to a given $\boldsymbol{s}$ is not fixed, 
%even with same state $\boldsymbol{s}$ as input, the result phase action $\boldsymbol{a}_i$ are not consistent, 
which makes phase prediction hard to converge. 
To have a more stable action for prediction, we use the action $a^r_0$ stored in the replay buffer of DQN as the target, which makes the phase prediction more stable and accurate. 
When the stored action $a^r_0$ is selected as the target, we decompose corresponding phase $P^r_0$ into the permissions $\boldsymbol{g}^r_0$ of traffic movements, and calculate the loss between recorded real permissions $\boldsymbol{g}^r_0$ and predicted permissions $\boldsymbol{g}^p_0$ as $L_p = \text{BinaryCrossEntropy}(\boldsymbol{g}^r_0, \boldsymbol{g}^p_0)$.

For learning approaching vehicle numbers $l_{0, y, z}$ prediction, we get vehicle numbers of every traffic movement from replay buffer of DQN, and learn $l_{0, y, z}$ through recorded results $l^r_{0, y, z}$. As actions saved in replay buffer may be different from the current actions, when calculating volume prediction loss, different from generating UniComm,
%when multiplying phase prediction and volume prediction, 
we use $\boldsymbol{g}^r_0$ instead of $\boldsymbol{g}_0$, i.e. $L_v=\text{MeanSquaredError}(\boldsymbol{g}^r_0 \cdot \boldsymbol{h}_0, l^r_{0, y, z})$.

%\baihua{I actually could not understand this paragraph, especially $g_i$/$g_j$.}
%Based on Eq.(\ref{eq:unicomm}), 
Based on how $c_1^t$ is derived, we also need to predict approaching vehicle number for next several intervals, which is rather challenging.
%there are some difficulties. 
Firstly, with off-policy learning, we can't really apply the selected action to the environment. 
Instead, we only learn from samples. 
Considering the phase prediction, while we can supervise the first phase taken by $A_0$, we, without interacting with the environment, don't know the real next state $s'\sim P(s, a)$, as the recorded $a^r_0$ may be different from $a_0$. 
The same problem applies to the prediction of $l_{0, y, z}$ too. 
%And the problem is also valid for $m_{i,j,k}$ prediction. 
Secondly, after multiple $\Delta t$ time intervals, the second assumption 
%\ref{ass:passroad} 
might become invalid, and it is hard to predict $l_{0, y, z}$ correctly with only $z_0$. 
As the result, we argue that as it is less useful to pass unsupervised and incorrect predictions, we only communicate with one time step prediction.

We list the pseudo code of UniComm in Algorithm~\ref{alg:unicomm}. 
As the process is the same for all intersections, for simplicity, 
we only present the code corresponding to one intersection. 
The input contains current intersection observation $z$, recorded next observation $z^r$, 
the set of traffic movements $T$ of the intersection, and turning direction $d$ of traffic movements. 
The algorithm takes observations $o^l$ belong to every traffic movement, 
and generates embeddings $\boldsymbol{h}$ based on $o^l$ (lines 4-5). 
The permission prediction uses $\boldsymbol{h}$ with self-Attention, linear transformation and Sigmoid function, where 
the linear transformation is to transform an embedding to a scalar (line 6). 
It next calculates the phase prediction loss with recorded phase permissions using BinaryCrossEntropy Loss (line 7).
Then, for every traffic movement $T_{a, x, b}\in T$, it accumulates embeddings to its corresponding outgoing lanes $\boldsymbol{h}^p_{x,b}$ and $\boldsymbol{h}^r_{x,b}$ with traffic movement embedding and permission respectively (lines 9-12). 
Finally, it uses linear transformation to predict the vehicle number of outgoing lanes $l$ and also volume prediction loss $L_v$ with MeanSquaredError Loss (lines 13-14). 
The algorithm outputs the estimated vehicle number $l$ for every outgoing lane of the intersection, and two prediction losses, including phase prediction loss $L_p$ and volume prediction loss $L_v$. Note that $z^r$ is only used to calculate $L_p$ and $L_v$.

\begin{algorithm}[tb]
\caption{UniComm Algorithm for an Intersection}
\label{alg:unicomm}
\textbf{Input}: intersection index $x$, observation $z$, recorded next observation $z^r$, traffic movements $T$, turning direction $d$ \\
\textbf{Parameter}: Traffic movement permission $\boldsymbol{g}$, traffic movement observation $o^l$, traffic movement embedding $\boldsymbol{h}$, 
out-lane prediction embedding $\boldsymbol{h}^p$, out-lane record embedding $\boldsymbol{h}^r$\\
\textbf{Output}: Estimated vehicle number $l$, phase loss $L_p$, volume loss $L_v$ 
\begin{algorithmic}[1] %[1] enables line numbers
\STATE $(\text{Vehicle number}\ n, \text{current phase}\ P) \gets z$ 
\STATE $(\text{Recorded approaching number}\ l^r, \text{phase}\ P^r) \gets z^r$ 
\STATE $\boldsymbol{g}^r \gets P^r$
\STATE $o^l_{a, x, b} \gets \{n_{a, x, b}, \boldsymbol{g}_{a, x, b}, d_{a, x, b}\}$
\STATE $\boldsymbol{h}_{a, x, b} \gets \text{Embedding}(o^l_{a, x, b})$
\STATE $\boldsymbol{g}^p_{a, x, b} \gets \text{Sigmoid}\left(\text{Self-Attention}(\boldsymbol{h}_{a, x, b})\right)$
\STATE Let $L_p \gets \text{BinaryCrossEntropy}(\boldsymbol{g}^r, \boldsymbol{g}^p)$

\STATE $\boldsymbol{h}^p \gets \boldsymbol{0}, \boldsymbol{h}^r \gets \boldsymbol{0}$
\FOR{$T_{a, x, b}$ \textbf{in} $T$}
\STATE $\boldsymbol{h}^p_{x, b} \gets \boldsymbol{h}^p_{x, b} + \boldsymbol{g}^p_{a, x, b}\cdot\boldsymbol{h}_{a, x, b}$
\STATE $\boldsymbol{h}^r_{x, b} \gets \boldsymbol{h}^r_{x, b} + \boldsymbol{g}^r_{a, x, b}\cdot\boldsymbol{h}_{a, x, b}$
\ENDFOR

\STATE $l \gets \text{Linear}(\boldsymbol{h}^p)$
\STATE $L_v \gets \text{MeanSquaredError}\left(\text{Linear}(\boldsymbol{h}^r), l^r\right)$

\STATE \textbf{return} $l$, $L_p$, $L_v$
\end{algorithmic}
\end{algorithm}

\subsection{UniLight}

Although UniComm is universal, its strength might not be fully achieved by existing methods, because they do not consider the importance of exchanged information. 
%
%as other methods don't consider the meaning of exchanged messages, they can't achieve best result with their original network structures. %
To make better use of predicted approaching vehicle numbers and other observations, we propose \emph{UniLight} to predict the Q-values. 

Take prediction of intersection $I_1$ for example. 
As we predict $l_{x, 1, y}$ based on traffic movements, UniLight splits average number of vehicles $n_{x,1,y}$, traffic movement permissions $\boldsymbol{g}_1$ and predictions $l_{x,1,y}$ into traffic movements $T_{x,1,y}$, and uses a fully-connected layer with ReLU to generate the hidden state $\boldsymbol{h}_1$. 
Next, considering one traffic phase $P$ that permits traffic movements $T_P$ among traffic movements $\mathcal{T}_1$ in $I_1$,
we split the hidden states $\boldsymbol{h}_1$ into two groups $G_1=\{h_p | T_p \in T_P\}$ and $G_2=\{h_x|h_x\in\boldsymbol{h}_1,T_x\notin T_P \}$, based on whether the corresponding movement is permitted by $P$ or not.  
As traffic movements in the same group will share the same permission in phase $P$, we consider that they can be treated equally and hence use the average of their hidden states to represent the group. 
Obviously, traffic movements in $G_1$ are more important than those in $G_2$, so we multiply the hidden state of $G_1$ with a greater weight to capture the importance.
Finally, we concatenate two hidden states of groups and a number representing whether current phase is $P$ through a fully-connected layer to predict the final Q-value. 

The pseudo code of UniLight is listed in Algorithm~\ref{alg:unilight}. 
Same as UniComm, we only present the code corresponding to one intersection. 
The input contains observation $z$, all possible phases, i.e. actions $\mathcal{A}$, 
traffic movements $T$ of the intersection, turning direction $d$ of traffic movements, 
and the estimated vehicle number $l$ for every traffic movement. 
It's worth to mention that UniComm estimates $l$ for every outgoing lane, but UniLight takes $l$ for incoming lanes as inputs. Consequently, we need to arrange the 
estimations in a proper manner such that the output of UniComm w.r.t. an intersection become the inputs of UniLight w.r.t. an adjacent intersection.
UniLight outputs Q-value estimation $Q(z, \mathcal{A})$
of current observation and all possible actions.
It takes the combination of observations and incoming vehicle estimation $o^l$ for every traffic movement, 
and generates the embeddings $\boldsymbol{h}$ for them (lines 3-4). For every action $a\in \mathcal{A}$, we split traffic movement embeddings into two groups $G_1$ and $G_2$ based on whether the corresponding traffic movement is permitted by $a$ or not, i.e., $G_1$ consists of the embeddings of all the movements in $T$ that are permitted by $a$ and $G_2$ consists of the embeddings of rest of the movements in $T$ that are stopped by $a$ (lines 6-9). We use the average of embeddings to represent a group (i.e., a group embedding), and multiply the group embedding with different weights. 
Obviously, traffic movements in $G_1$ are more important than the rest traffic movements in $G_2$ for action $a$, so $w_1$ is much larger than $w_2$.
Finally, we concatenate group embeddings with current phase $P$, and use a linear transformation to predict $Q(z, a)$ (line 10).

\begin{algorithm}[tb]
\caption{UniLight Algorithm for an Intersection}
\label{alg:unilight}
\textbf{Input}: Intersection index $x$, observation $z$, possible phases $\mathcal{A}$, traffic movements $T$, turning direction $d$, estimated vehicle number $l$\\
\textbf{Parameter}: Traffic movement permission $\boldsymbol{g}$, traffic movement observation $o_l$, traffic movement embedding $\boldsymbol{h}$, action phase $a$ \\
\textbf{Output}: Q-values $Q(z, \mathcal{A})$ 
\begin{algorithmic}[1] %[1] enables line numbers
\STATE $(\text{Vehicle number}\ n, \text{current phase}\ P) \gets z$ 
\STATE $\boldsymbol{g} \gets P$
\STATE $o^l_{a, x, b} \gets \{n_{a, x, b}, \boldsymbol{g}_{a, x, b}, d_{a, x, b}, l_{a, x, b}\}$
\STATE $\boldsymbol{h}_{a, x, b} \gets \text{Embedding}(o^l_{a, x, b})$
\FOR{$a$ \textbf{in} $\mathcal{A}$}
\STATE $g^a \gets a$
\STATE $T_a \gets \{T_i \in T | i: \boldsymbol{g}^a_i \text{~is permitted} \}$
\STATE  $G_1 \gets \{\boldsymbol{h}_i \in \boldsymbol{h} | i: T_i \in T_a\}$ 
\STATE  $G_2 \gets \boldsymbol{h}\backslash G_1$
\STATE  $Q(z, a) \gets \text{Linear}\left(w_1\overline{G_1} \oplus w_2\overline{G_2} \oplus P\right)$
\ENDFOR
\STATE \textbf{return} $Q(z,\mathcal{A})$
\end{algorithmic}
\end{algorithm}

\section{Experiments}

In this section, we first explain the detailed experimental setup, including implementaion details, datasets used, competitors evaluated, and performance metrics employed; we then report the experimental results. 

\subsection{Experimental Setup}

\subsubsection{Implementation details}

We conduct our experiments on the microscopic traffic simulator CityFlow~\cite{zhang2019cityflow}. 
We have made some modifications to CityFlow to support the collection of structural data from the intersections and the roads.
In the experiment, the action time interval $\Delta t=10s$, and the all-red phase is $5s$. 
We run the experiments on a cloud platform with 2 virtual Intel 8269CY CPU core and 4GB memory. 
We train all the methods, including newly proposed ones and the existing ones selected for comparison, with $240,000$ frames. We run all the experiments 4 times, and select the best one which will be then tested 10 times with its average performance and its standard deviation reported in our paper.

Most experiments are run on Unbuntu 20.04, PyTorch 1.7.1 with CPU only. 
A small part of experiments are run on GPU machines.
We list the parameter settings in the following and please refer to the source codes for more detailed implementations.
%please refer to the code.

For Double-Dueling DQN, we use discount factor $\gamma=0.8$ and 5-step learning.
The replay buffer size is 8000. When replay buffer is fulfilled, 
the model is trained after every step with batch size 30, 
and the target network is updated every 5 steps. 
For $\epsilon$-greedy strategy, the initial $\epsilon=0.9$, the minimum $\epsilon=0.02$,
and $\epsilon$ is decreasing linearly in the first 30\% training steps.

%As the lane number of different roads are not same, 
As roads have different number of lanes
and different traffic movements have various lane numbers, 
the average number of vehicles of one traffic movement $n$ in the observation $z$ is divided by its lane number.
In UniComm, the dimension of lane embedding $\boldsymbol{h}[i]$ is 32,
turning direction $d[i]$ is represented by an embedding with dimension 2, and 
Self-Attention used in phase prediction has one head. 
In UniLight, the dimension of lane embedding $h[i]$ is 32, 
and the weights for two groups are $w_1=5$ and $w_2=1$ respectively. 

%\subsection{Datasets}

\subsubsection{Datasets} 
We use the real-world datasets from four different cities, including Hangzhou (HZ), Jinan (JN), and Shanghai (SH) from China, and New York (NY) from USA. 
The HZ, JN, and NY datasets are publicly available~\cite{wei2019colight}, and widely used in many related studies. 
%
%However, despite of 
However, the simulation time of these three datasets is relatively short, and hence it is hard to test the performance of a signal controlling method, especially during the rush hour. 
We also notice that roads in these three datasets contain exactly three lanes, i.e., one lane for every turning direction. 
In those public datasets, the length of roads sharing the same direction is a constant (e.g., all the vertical roads share the same road length), 
which is different from the reality. 
Meanwhile, although the traffic data of public datasets is based on real traffic flow, they adopt a fixed probability distribution, as it is not possible to track all the vehicle trajectories. 
For example, they assume 10\% of traffic flow turn left, 60\% go straight, and 30\% turn right to simulate vehicle trajectories, which is very unrealistic.

Motivated by the limitations of public datasets, we try to develop new datasets that can simulate the urban traffic in a more realistic manner. 
As taxis perform an important role in urban traffic, we utilize substantial taxi trajectories to reproduce the traffic flow for an area closer to the reality. 
Shanghai taxi dataset contains $11,861,593$ trajectories generated by $13,767$ taxis in the whole month of April, 2015.
Directly using taxi trajectories in one day will face the data sparsity problem. 
To overcome data sparsity, we squash all trajectories into one day to decrease the variance and meanwhile balance the number of vehicle routes based on the traffic volume from official data to truthfully recover the traffic of one area\footnote{In 2015, there were 2.5 million registered vehicles in Shanghai. There are close to 12 million trajectories in our Shanghai taxi dataset, corresponding to on average 4 to 5 trips per vehicle in a day.}. 
We choose two transportation hubs, which contain roads of different levels, i.e. roads have different number of lanes, denoted as SH$_1$ and SH$_2$ respectively. In our datasets, the two road segments belonging to the same road but having opposite directions have the same number of lanes. 

\begin{figure}
    \centering
    \includegraphics[width=24mm]{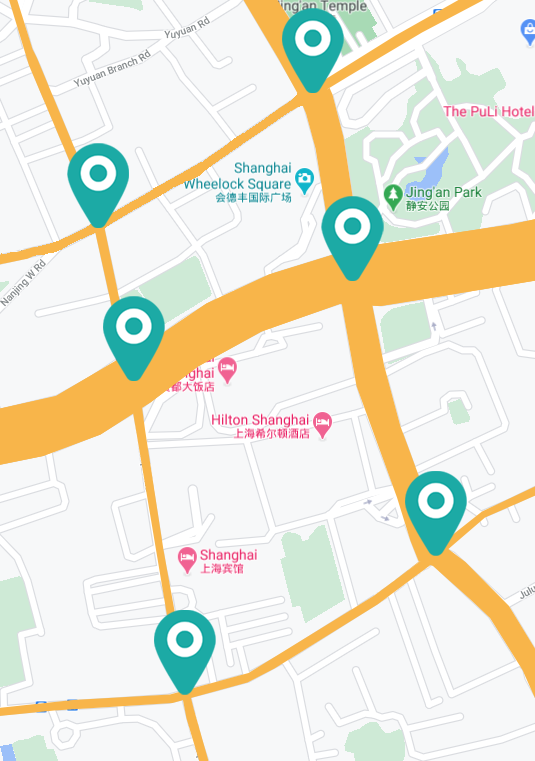}
    \includegraphics[width=47mm]{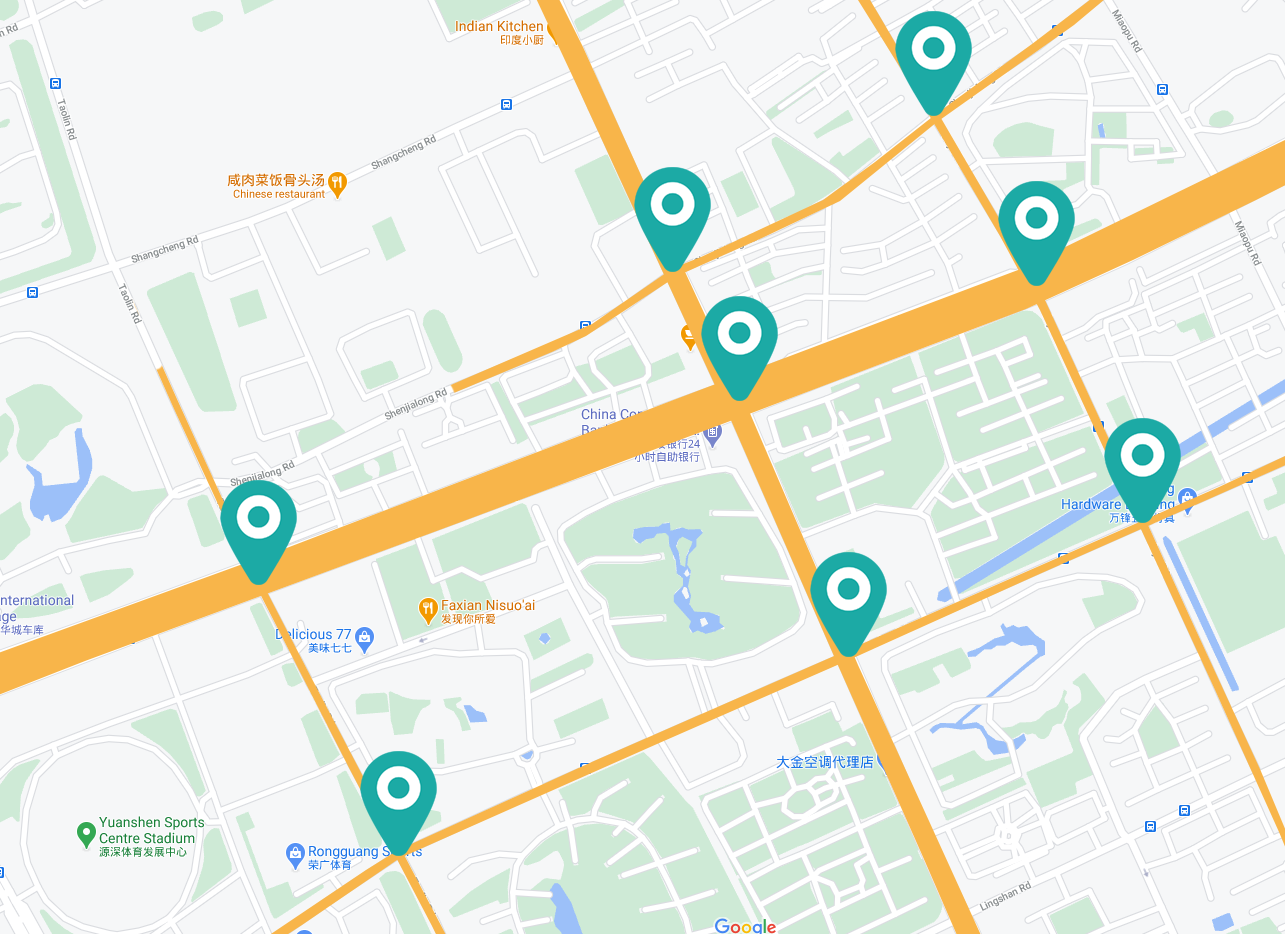}
    %\vspace{-0.1in}
	\caption{Real road networks of two Shanghai datasets with 6 and 8 intersections respectively. Roads with more lanes are colored with wider lines.
	%\baihua{Do we need to explain the sizes of those two hubs?}
	}
	\label{fig:shanghaimap}
    %\vspace{-5mm}
\end{figure}

Figure~\ref{fig:shanghaimap} visualizes those two hubs. 
The first hub SH$_1$ contains six intersections, which are marked by green icons. 
The lane numbers of the three horizontal roads are 3, 7, and 3 respectively, and 
the lane numbers of the two vertical roads are 3 and 4 respectively. For the second hub SH$_2$, it contains eight intersections.  The three horizontal roads have 3, 5, and 3 lanes respectively, while all the vertical roads have 3 lanes. 
As stated previously, for all the roads in these two hubs, they share the same number of lanes on both directions. For example, the very wide horizontal road in SH$_2$ corresponds a bidirectional road in reality, and it is represented by two roads in our simulation. Those two roads have opposite directions, one from the east to the west and the other from the west to the east, but they have the same number of lanes. 
In reality, some small roads only contain one or two lanes, and some turning directions share one lane, 
e.g., in a two-lane road, the left lane can be used to turn left, 
and the right lane can be used to go-straight or turn right.
However, lanes in CityFlow cannot share multiple turning directions as CityFlow allows each lane to have exact one turning direction. 
Consequently, we have to add lanes to those roads such that all the turning directions allowed in reality can be supported too in our experiments. 
This explains why in the second hub, 
the vertical road located in the middle looks wider than the other two vertical roads but it shares the same number of lanes as the other two vertical roads.

We show the statistics of all five datasets in Table~\ref{tab:datastatistic}. 
The average number of vehicles in 5 minutes is counted based on every single intersection in the dataset.
Compared with existing datasets, the roads in our newly generated datasets SH$_1$ and SH$_2$ are more realistic. They have  different number of lanes and different length, much wider time span,
and the vehicle trajectories are more realistic and dynamic.

\linespread{1.2}
\begin{table*}[htb]
\centering
\begin{tabular}{l|ccc|cc}
\hline
\cline{2-6}
    & HZ      & JN      & NY      & SH$_1$    & SH$_2$     \\ \hline
\# of Intersections         & 16      & 12      & 48      & 6      & 8       \\
Road Length (metre)         & 600/800     & 400/800     & 100/350     & 62$\sim$612 & 102$\sim$706 \\
\# of Lanes         & 3       & 3       & 3       & 3$\sim$7    & 3$\sim$5     \\
Time Span (second)    & 3600   & 3600   & 3600   & 86400 & 86400  \\
\# of vehicle        & 2983    & 6295    & 2824    & 22313  & 15741   \\
Average \# of vehicles in 5 minutes & 13$\sim$21 & 21$\sim$57 & 5$\sim$6 & 3$\sim$76 & 2$\sim$52  \\
\hline
\end{tabular}
\caption{Datasets Statistics}
\label{tab:datastatistic}
\end{table*}

\subsubsection{Competitors}
To evaluate the effectiveness of newly proposed \textbf{UniComm} and \textbf{UniLight}, we implement five conventional TSC methods and six representative reinforcement learning M-TSC methods as competitors, which are listed below. 
%These methods selected for comparison are summarized below. 

(1) \textbf{SOTL}~\cite{Cools2013}, a S-TSC method 
%that switches traffic signals 
based on current vehicle numbers on every traffic movement.
%    \item 
%
%\noindent    

(2) \textbf{MaxPressure}~\cite{varaiya2013max}, a M-TSC method that balances the vehicle numbers between two neighboring intersections.

(3) \textbf{MaxBand}~\cite{little1981maxband}
%, a conventional method 
that maximizes the green wave time for both directions of a road.

(4) \textbf{TFP-TCC}~\cite{DBLP:conf/icaci/JiangHC21} that predicts the traffic flow and uses traffic congestion control methods based on future traffic condition. 

(5) \textbf{MARLIN}~\cite{el2013multiagent} that uses Q-learning to learn joint actions for agents. 
%with their neighbors;
%\qize{for urban and marlin, it is implemented by teammate, I will update method description in Jan 12.}
%    \item 
%
%\noindent    

(6) \textbf{MA-A2C}~\cite{DBLP:conf/icml/MnihBMGLHSK16}, a general reinforcement learning method with actor-critic structure and multiple agents.

(7) \textbf{MA-PPO}~\cite{DBLP:journals/corr/SchulmanWDRK17}, a popular policy based method, which improves MA-A2C with proximal policy optimization to stable the training process.

(8) \textbf{PressLight}~\cite{wei2019presslight}, a 
reinforcement learning based method motivated by MaxPressure,
which modifies the reinforcement learning RL design to use pressure as the metric.
%\item 
%
%\noindent

(9) \textbf{CoLight}~\cite{wei2019colight} that uses graph attentional network to communicate between neighboring intersections with different weights. 
%\item 
%
%\noindent

(10) \textbf{MPLight}~\cite{chen2020toward}, a state-of-the-art M-TSC method that combines FRAP~\cite{zheng2019learning} and PressLight, and is one of the state-of-the-art methods in M-TSC. 

(11) \textbf{AttendLight}~\cite{DBLP:conf/nips/OroojlooyNHS20} that uses attention and LSTM to select best actions. 
%\baihua{What is FRAP?}\qize{add ref when back to lab}
%\item 

%\noindent
%\textbf{UniComm}. Our proposed communication form. We adapt it to existing reinforcement learning methods to evaluate the performance. 
%\item 
%
%\textbf{UniLight}. Our proposed traffic signal control method which is designed to make full use of UniComm, as well as observations. The normal version means we never communicate between agents. 
    %\item \textbf{UniLight-C}. We \emph{remove the communication} from UniComm from UniLight in this version.
    %\item \textbf{UniLight-B}. Instead of sending the prediction of approaching vehicle numbers, we \emph{send a 32-dimension embedding} of the intersection as most methods do in this version. 
%\end{itemize}

\subsubsection{Performance Metrics}

To prove the effectiveness of our proposed methods, we have performed a comprehensive experimental study to evaluate the performance of my methods and all the competitors. 
Following existing studies~\cite{DBLP:conf/nips/OroojlooyNHS20,chen2020toward}, we consider in total four different performance metrics, including average travel time, average delay, average wait time and throughput. The meanings of different metrics are listed below.

\begin{itemize}
    \item \textbf{Travel time.} The travel time of a vehicle is the time it used between entering and leaving the environment.
    \item \textbf{Delay.} The delay of a vehicle is its travel time minus the expected travel time, i.e., the time used to travel when there is no other vehicles and traffic signals are always green.
    \item \textbf{Wait time.} The wait time is defined as the time a vehicle is waiting, i.e., its speed is less than a threshold. In our experiments, the threshold is set to 0.1m/s.
    \item \textbf{Throughput.} The throughput is the number of vehicles that have completed their trajectories before the simulation stops.
\end{itemize}

%\subsection{Evaluation Metric}
%
%Following existing studies, to evaluate the performance realistically, we use the \emph{average travel time} of all vehicles as the evaluation metric.

\subsection{Evaluation Results}

\linespread{1.2}

\begin{table}[htb]
\resizebox{0.48\textwidth}{!}{%
\begin{tabular}{cc|c|c|c|c|c}
\multicolumn{2}{c|}{Datasets}                          & JN   & HZ & NY & SH$_1$ & SH$_2$ \\ \hline
\multicolumn{1}{c|}{\multirow{3}{*}{\rotatebox[origin=c]{90}{UniLight}}} & No Com.       & 335.85 & 324.24 & 186.85 & 2326.29 & 209.89  \\
\multicolumn{1}{c|}{}                   & Hidden State & 330.99 & 323.88 & 180.99 & 1874.11 & 224.55  \\
\multicolumn{1}{c|}{}                   & UniComm      & \textbf{325.47} & \textbf{323.01} & \textbf{180.72} & \textbf{159.88} & \textbf{208.06} 
\end{tabular}%
}
\caption{Compare UniComm with hidden state in average travel time.}
\label{tab:unicommvshs}
\end{table}

\begin{figure*}[tb]
    \centering
    \includegraphics[width=56mm]{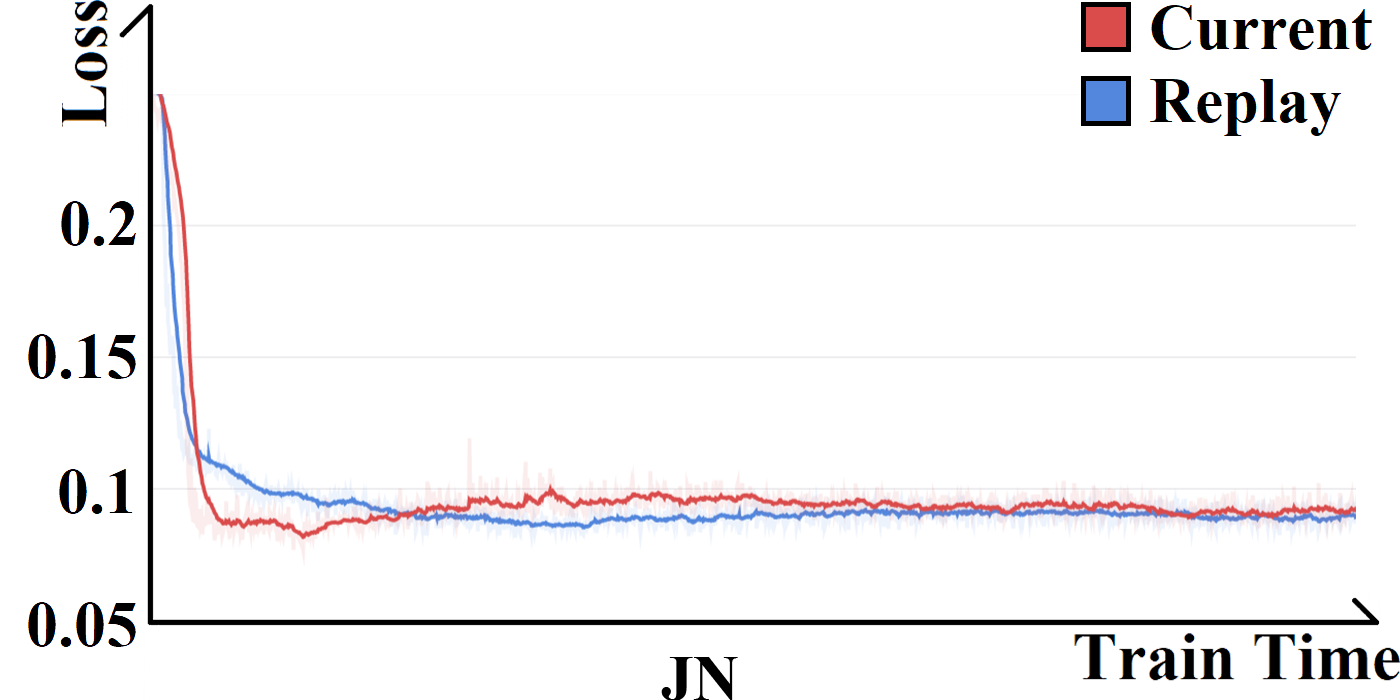}
    \includegraphics[width=56mm]{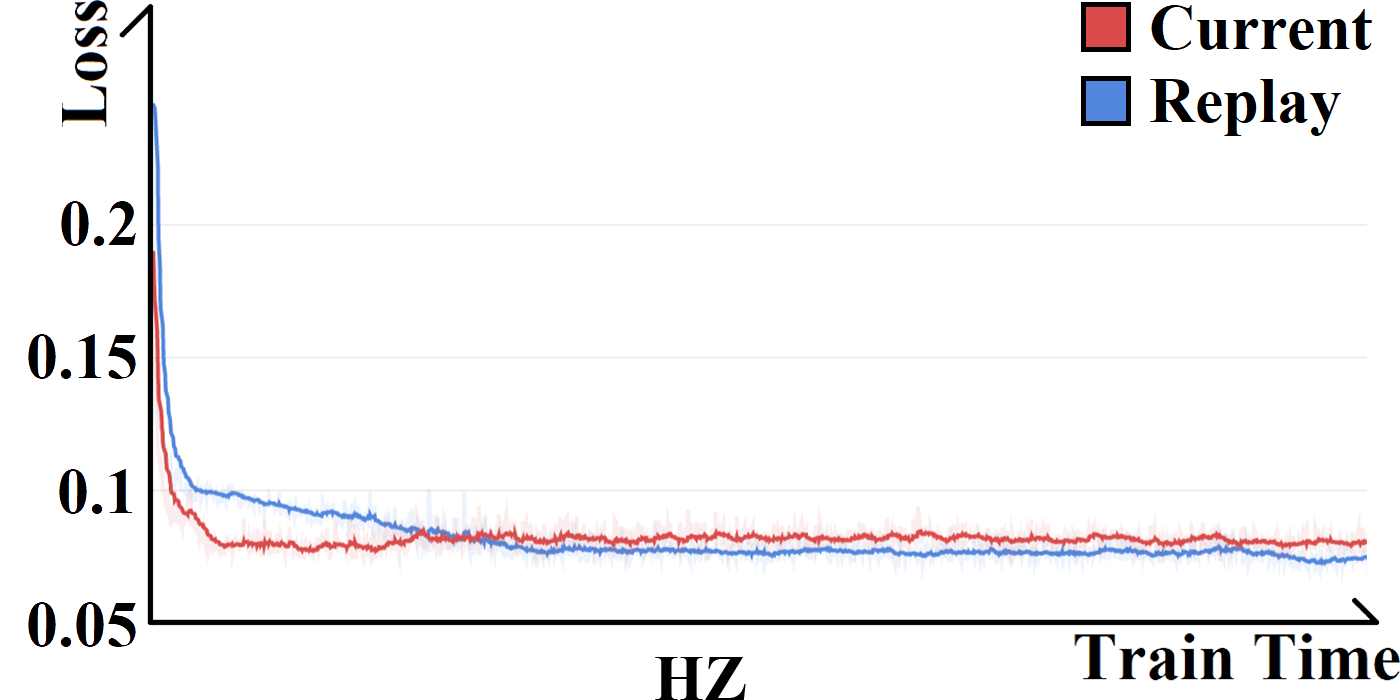}
    \includegraphics[width=56mm]{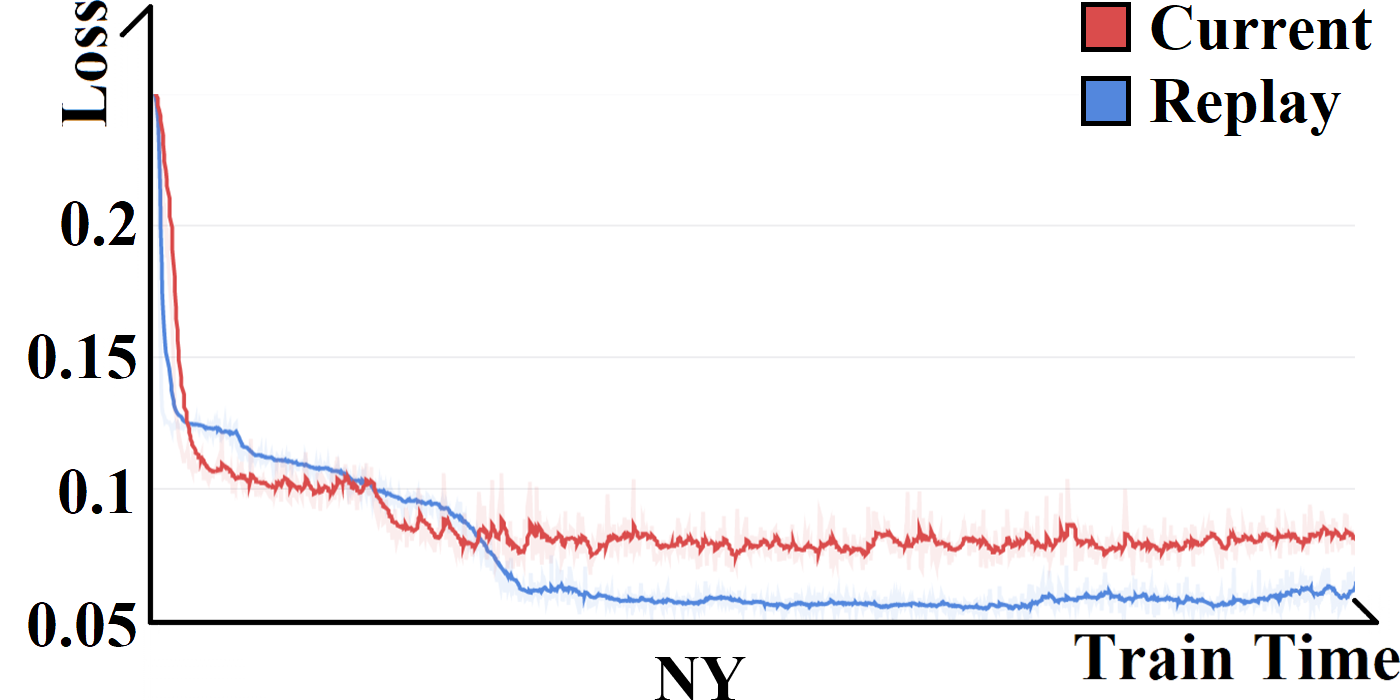}
    \includegraphics[width=56mm]{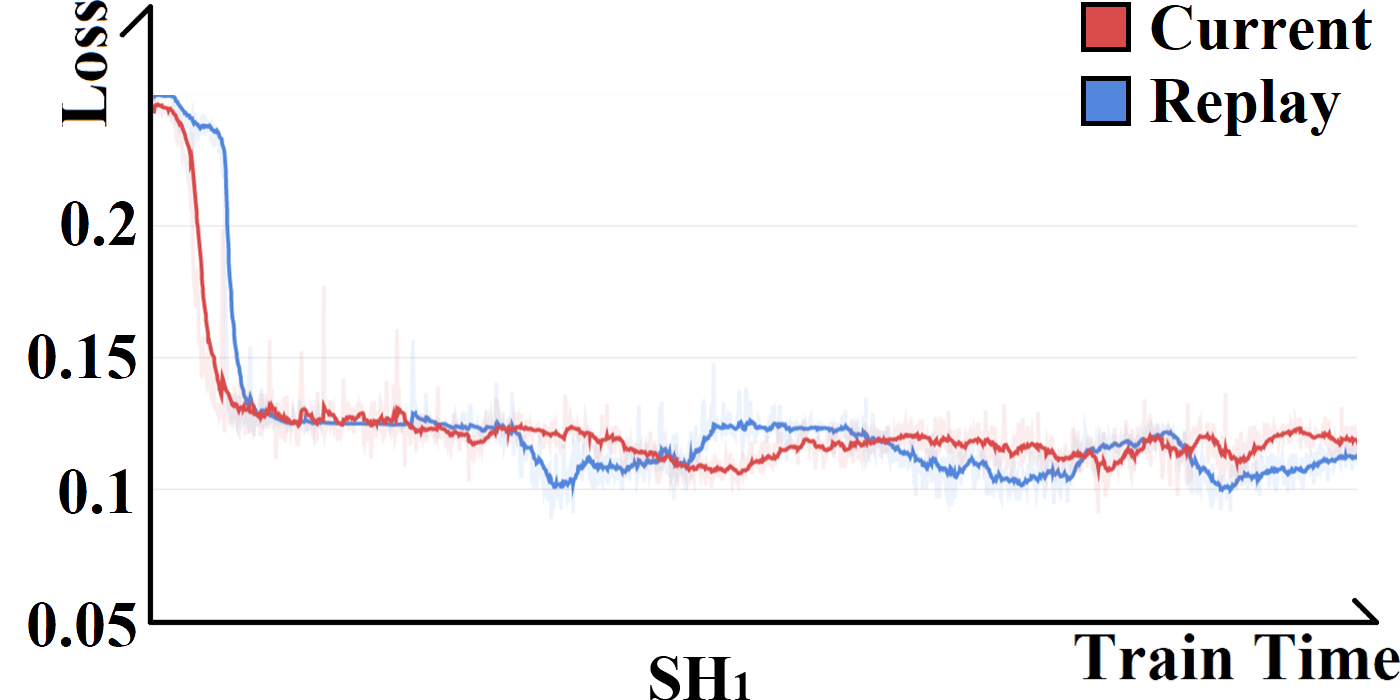}
    \includegraphics[width=56mm]{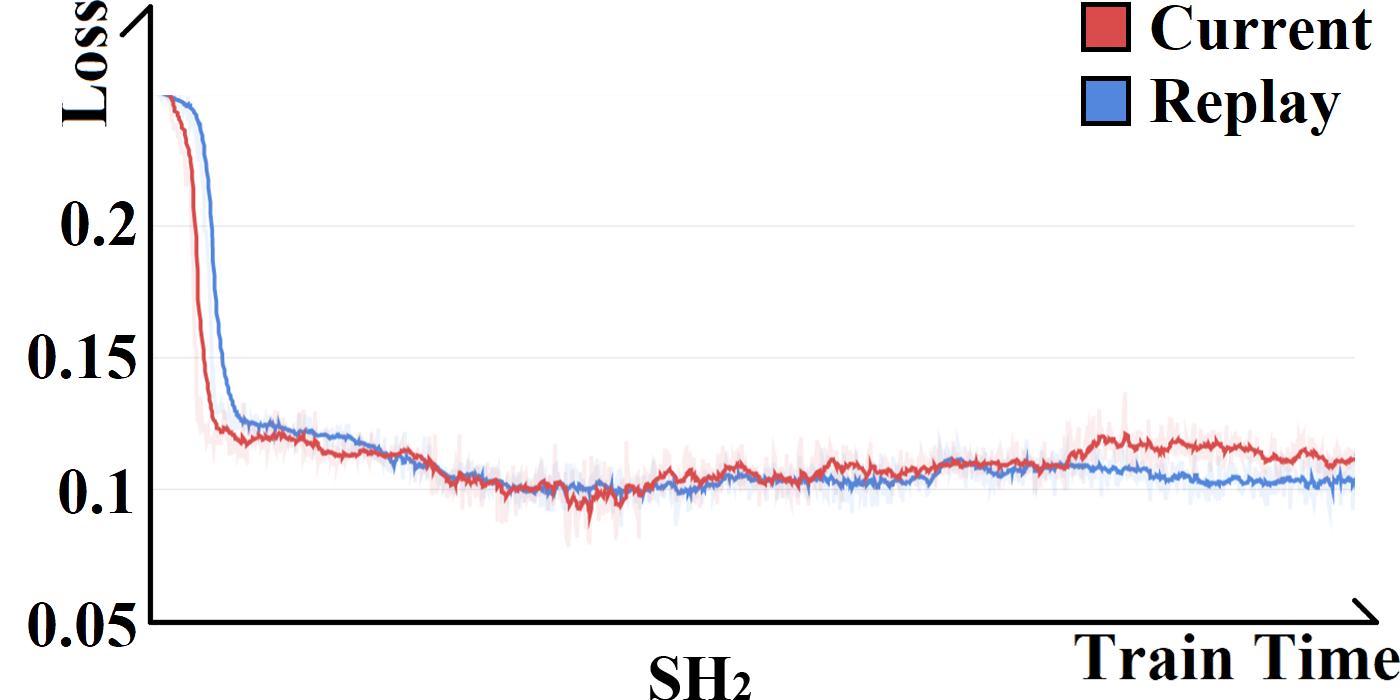}
    %\vspace{-0.1in}
	\caption{Phase prediction loss of different phase prediction target.}
	\label{fig:phaselossall}
    %\vspace{-5mm}
\end{figure*}

\begin{table*}[htb]
\resizebox{\textwidth}{!}{%
\begin{tabular}{ccc||c|c|c|c|c}
\multicolumn{3}{c||}{Datasets} & {JN} & {HZ} & {NY} & {SH$_1$} & {SH$_2$} \\
\hline
\multirow{5}{*}{\rotatebox[origin=c]{90}{Traditional}} & \multicolumn{1}{c|}{\multirow{5}{*}{\rotatebox[origin=c]{90}{Algorithms}}} &
SOTL &
420.70 $\pm$ 0.00 &451.62 $\pm$ 0.00 &1137.29 $\pm$ 0.00 &2362.73 $\pm$ 216.82 &2084.93 $\pm$ 82.83 
\\
& \multicolumn{1}{c|}{} & MaxPressure &
434.75 $\pm$ 0.00 &408.47 $\pm$ 0.00 &243.59 $\pm$ 0.00 &4843.07 $\pm$ 93.82 &788.55 $\pm$ 50.06 
\\
& \multicolumn{1}{c|}{} &
MaxBand &
378.45 $\pm$ 2.06 &458.00 $\pm$ 1.41 &1038.67 $\pm$ 12.81 &5675.98 $\pm$ 122.45 &4501.39 $\pm$ 85.04 
\\
& \multicolumn{1}{c|}{} &
TFP-TCC &
362.70 $\pm$ 5.30 &400.24 $\pm$ 1.42 &1216.77 $\pm$ 21.27 &2403.66 $\pm$ 878.66 &1380.05 $\pm$ 146.18 
\\
& \multicolumn{1}{c|}{} & MARLIN &
409.27 $\pm$ 0.00 &388.00 $\pm$ 0.03 &1274.23 $\pm$ 0.00 &6623.83 $\pm$ 507.52 &5373.53 $\pm$ 101.74 
\\
\hline
\multirow{7}{*}{\rotatebox[origin=c]{90}{DRL}} & \multicolumn{1}{c|}{\multirow{7}{*}{\rotatebox[origin=c]{90}{Algorithms}}}
& MA-A2C &
355.29 $\pm$ 1.03 &353.28 $\pm$ 0.82 &906.71 $\pm$ 28.96 &4787.83 $\pm$ 362.12 &431.53 $\pm$ 41.01 
\\
& \multicolumn{1}{c|}{} & MA-PPO &
460.18 $\pm$ 20.71 &352.64 $\pm$ 1.18 &923.82 $\pm$ 22.77 &3650.83 $\pm$ 171.07 &2026.71 $\pm$ 597.83 
\\
& \multicolumn{1}{c|}{} & PressLight &
335.93 $\pm$ 0.00 &338.41 $\pm$ 0.00 &1363.31 $\pm$ 0.00 &5114.78 $\pm$ 647.92 &322.48 $\pm$ 3.84 
\\
& \multicolumn{1}{c|}{}
& CoLight &
329.67 $\pm$ 0.00 &340.36 $\pm$ 0.00 &244.57 $\pm$ 0.00 &7861.59 $\pm$ 69.86 &4438.90 $\pm$ 260.66 
\\
& \multicolumn{1}{c|}{}
& MPLight &
383.95 $\pm$ 0.00 &334.04 $\pm$ 0.00 &646.94 $\pm$ 0.00 &7091.79 $\pm$ 22.97 &433.92 $\pm$ 23.26 
\\
& \multicolumn{1}{c|}{}
& AttendLight &
361.94 $\pm$ 2.85 &339.45 $\pm$ 0.82 &1376.81 $\pm$ 16.41 &4700.22 $\pm$ 87.50 &2763.66 $\pm$ 425.19 
\\
\cline{3-8}
& \multicolumn{1}{c|}{}
& UniLight &
335.85 $\pm$ 0.00 &324.24 $\pm$ 0.00 &186.85 $\pm$ 0.00 &2326.29 $\pm$ 242.90 &209.89 $\pm$ 21.70 
\\
\hline
\multirow{7}{*}{\rotatebox[origin=c]{90}{With}} & \multicolumn{1}{c|}{\multirow{7}{*}{\rotatebox[origin=c]{90}{UniComm}}}
& MA-A2C &
332.80 $\pm$ 1.71 &349.93 $\pm$ 1.09 &834.65 $\pm$ 38.09 &4018.67 $\pm$ 319.11 &303.69 $\pm$ 6.13 
\\
& \multicolumn{1}{c|}{} & MA-PPO &
331.96 $\pm$ 1.34 &349.82 $\pm$ 1.21 &847.49 $\pm$ 30.88 &3806.77 $\pm$ 194.88 &290.99 $\pm$ 4.23 
\\
& \multicolumn{1}{c|}{} & PressLight &
\textbf{317.72 $\pm$ 0.00} &330.28 $\pm$ 0.00 &1152.76 $\pm$ 0.00 &6200.91 $\pm$ 529.39 &549.56 $\pm$ 51.20 
\\
& \multicolumn{1}{c|}{}
& CoLight &
318.93 $\pm$ 0.00 &336.66 $\pm$ 0.00 &291.40 $\pm$ 0.00 &7612.02 $\pm$ 271.91 &1422.99 $\pm$ 633.09 
\\
& \multicolumn{1}{c|}{}
& MPLight &
336.29 $\pm$ 0.00 &329.57 $\pm$ 0.00 &193.21 $\pm$ 0.00 &5095.34 $\pm$ 224.42 &542.82 $\pm$ 119.56 
\\
& \multicolumn{1}{c|}{}
& AttendLight &
363.41 $\pm$ 3.79 &330.38 $\pm$ 1.08 &608.12 $\pm$ 38.59 &4825.83 $\pm$ 249.90 &2915.35 $\pm$ 757.95 
\\
\cline{3-8}
& \multicolumn{1}{c|}{}
& UniLight &
325.47 $\pm$ 0.00 &\textbf{323.01 $\pm$ 0.00} &\textbf{180.72 $\pm$ 0.00} &\textbf{159.88 $\pm$ 1.87} &\textbf{208.06 $\pm$ 0.88 }
\end{tabular}%

}
\caption{Evaluation result of average travel time (seconds).}
\label{tab:travel}
\end{table*}

\iffalse
\begin{table*}[htb]
\resizebox{\textwidth}{!}{%

}
\caption{Evaluation result of average travel time from all 4 runs.}
\label{tab:travelall}
\end{table*}
\fi

\begin{table*}[htb]
\resizebox{\textwidth}{!}{%
\begin{tabular}{ccc||c|c|c|c|c}
\multicolumn{3}{c||}{Datasets} & {JN} & {HZ} & {NY} & {SH$_1$} & {SH$_2$} \\
\hline
\multirow{5}{*}{\rotatebox[origin=c]{90}{Traditional}} & \multicolumn{1}{c|}{\multirow{5}{*}{\rotatebox[origin=c]{90}{Algorithms}}} &
SOTL &
260.69 $\pm$ 0.17 &232.33 $\pm$ 0.04 &1079.57 $\pm$ 0.00 &2340.80 $\pm$ 217.95 &2047.33 $\pm$ 84.18 
\\
& \multicolumn{1}{c|}{} & MaxPressure &
269.30 $\pm$ 0.04 &173.03 $\pm$ 0.00 &120.52 $\pm$ 0.00 &4827.94 $\pm$ 94.16 &738.10 $\pm$ 50.68 
\\
& \multicolumn{1}{c|}{} &
MaxBand &
216.83 $\pm$ 2.40 &237.20 $\pm$ 1.50 &973.48 $\pm$ 13.43 &5663.45 $\pm$ 122.96 &4479.86 $\pm$ 85.46 
\\
& \multicolumn{1}{c|}{} &
TFP-TCC &
199.80 $\pm$ 5.11 &182.46 $\pm$ 1.65 &1161.35 $\pm$ 22.28 &2380.09 $\pm$ 884.42 &1334.86 $\pm$ 147.56 
\\
& \multicolumn{1}{c|}{} & MARLIN &
247.59 $\pm$ 0.04 &158.49 $\pm$ 0.08 &1220.23 $\pm$ 0.00 &6614.58 $\pm$ 508.83 &5354.42 $\pm$ 101.87 
\\
\hline
\multirow{7}{*}{\rotatebox[origin=c]{90}{DRL}} & \multicolumn{1}{c|}{\multirow{7}{*}{\rotatebox[origin=c]{90}{Algorithms}}}
& MA-A2C &
190.84 $\pm$ 1.29 &121.30 $\pm$ 1.00 &834.03 $\pm$ 30.93 &4773.34 $\pm$ 363.34 &378.63 $\pm$ 41.98 
\\
& \multicolumn{1}{c|}{} & MA-PPO &
305.78 $\pm$ 22.30 &120.53 $\pm$ 1.51 &852.49 $\pm$ 24.34 &3631.40 $\pm$ 171.77 &1984.16 $\pm$ 603.71 
\\
& \multicolumn{1}{c|}{} & PressLight &
169.30 $\pm$ 0.06 &100.17 $\pm$ 0.00 &1314.57 $\pm$ 0.00 &5100.96 $\pm$ 649.87 &267.35 $\pm$ 3.72 
\\
& \multicolumn{1}{c|}{}
& CoLight &
162.71 $\pm$ 0.07 &98.84 $\pm$ 0.04 &125.32 $\pm$ 0.00 &7854.45 $\pm$ 70.14 &4411.89 $\pm$ 263.23 
\\
& \multicolumn{1}{c|}{}
& MPLight &
218.41 $\pm$ 0.03 &88.68 $\pm$ 0.00 &554.87 $\pm$ 0.00 &7083.68 $\pm$ 23.03 &379.99 $\pm$ 23.65 
\\
& \multicolumn{1}{c|}{}
& AttendLight &
197.32 $\pm$ 2.62 &103.16 $\pm$ 1.58 &1329.94 $\pm$ 17.37 &4684.92 $\pm$ 87.76 &2731.29 $\pm$ 428.06 
\\
\cline{3-8}
& \multicolumn{1}{c|}{}
& UniLight &
166.49 $\pm$ 0.01 &74.93 $\pm$ 0.00 &55.85 $\pm$ 0.00 &2302.12 $\pm$ 244.06 &153.11 $\pm$ 22.65 
\\
\hline
\multirow{7}{*}{\rotatebox[origin=c]{90}{With}} & \multicolumn{1}{c|}{\multirow{7}{*}{\rotatebox[origin=c]{90}{UniComm}}}
& MA-A2C &
166.91 $\pm$ 1.73 &117.20 $\pm$ 1.63 &757.19 $\pm$ 40.58 &4000.45 $\pm$ 320.40 &249.05 $\pm$ 6.21 
\\
& \multicolumn{1}{c|}{} & MA-PPO &
165.65 $\pm$ 1.38 &116.75 $\pm$ 2.20 &771.26 $\pm$ 32.83 &3788.19 $\pm$ 195.69 &236.43 $\pm$ 4.39 
\\
& \multicolumn{1}{c|}{} & PressLight &
\textbf{149.92 $\pm$ 0.05} &86.35 $\pm$ 0.00 &1096.51 $\pm$ 0.00 &6189.67 $\pm$ 530.99 &497.70 $\pm$ 52.01 
\\
& \multicolumn{1}{c|}{}
& CoLight &
151.00 $\pm$ 0.12 &94.24 $\pm$ 0.00 &176.94 $\pm$ 0.00 &7605.18 $\pm$ 272.52 &1379.88 $\pm$ 637.67 
\\
& \multicolumn{1}{c|}{}
& MPLight &
166.26 $\pm$ 0.07 &81.76 $\pm$ 0.00 &63.52 $\pm$ 0.00 &5081.75 $\pm$ 225.04 &491.35 $\pm$ 121.36 
\\
& \multicolumn{1}{c|}{}
& AttendLight &
197.60 $\pm$ 4.23 &86.03 $\pm$ 1.76 &507.25 $\pm$ 39.91 &4811.16 $\pm$ 250.93 &2884.94 $\pm$ 763.61 
\\
\cline{3-8}
& \multicolumn{1}{c|}{}
& UniLight &
156.33 $\pm$ 0.04 &\textbf{72.63 $\pm$ 0.00} &\textbf{48.47 $\pm$ 0.00} &\textbf{119.99 $\pm$ 1.84} &\textbf{151.63 $\pm$ 0.93} 
\end{tabular}%

}
\caption{Evaluation result of average delay (seconds).}
\label{tab:delay}
\end{table*}

\begin{table*}[htb]
\resizebox{\textwidth}{!}{%
\begin{tabular}{ccc||c|c|c|c|c}
\multicolumn{3}{c||}{Datasets} & {JN} & {HZ} & {NY} & {SH$_1$} & {SH$_2$} \\
\hline
\multirow{5}{*}{\rotatebox[origin=c]{90}{Traditional}} & \multicolumn{1}{c|}{\multirow{5}{*}{\rotatebox[origin=c]{90}{Algorithms}}} &
SOTL &
181.40 $\pm$ 0.00 &149.48 $\pm$ 0.00 &1057.95 $\pm$ 0.00 &2309.50 $\pm$ 220.12 &1988.22 $\pm$ 87.12 
\\
& \multicolumn{1}{c|}{} & MaxPressure &
197.85 $\pm$ 0.00 &117.95 $\pm$ 0.00 &66.44 $\pm$ 0.00 &4813.49 $\pm$ 94.59 &679.05 $\pm$ 53.07 
\\
& \multicolumn{1}{c|}{} &
MaxBand &
138.97 $\pm$ 1.95 &158.44 $\pm$ 1.42 &946.18 $\pm$ 14.03 &5646.98 $\pm$ 124.50 &4448.32 $\pm$ 86.18 
\\
& \multicolumn{1}{c|}{} &
TFP-TCC &
124.67 $\pm$ 4.92 &105.31 $\pm$ 1.48 &1142.30 $\pm$ 22.99 &2346.91 $\pm$ 894.53 &1268.55 $\pm$ 148.46 
\\
& \multicolumn{1}{c|}{} & MARLIN &
175.82 $\pm$ 0.00 &92.56 $\pm$ 0.03 &1202.31 $\pm$ 0.00 &6602.23 $\pm$ 510.35 &5329.60 $\pm$ 104.08 
\\
\hline
\multirow{7}{*}{\rotatebox[origin=c]{90}{DRL}} & \multicolumn{1}{c|}{\multirow{7}{*}{\rotatebox[origin=c]{90}{Algorithms}}}
& MA-A2C &
110.93 $\pm$ 1.06 &54.30 $\pm$ 0.67 &798.16 $\pm$ 32.82 &4756.17 $\pm$ 365.55 &298.59 $\pm$ 41.97 
\\
& \multicolumn{1}{c|}{} & MA-PPO &
234.25 $\pm$ 25.70 &53.51 $\pm$ 1.12 &817.35 $\pm$ 25.36 &3607.33 $\pm$ 172.69 &1939.55 $\pm$ 609.96 
\\
& \multicolumn{1}{c|}{} & PressLight &
94.89 $\pm$ 0.00 &41.62 $\pm$ 0.00 &1292.71 $\pm$ 0.00 &5083.04 $\pm$ 652.50 &197.60 $\pm$ 3.86 
\\
& \multicolumn{1}{c|}{}
& CoLight &
90.75 $\pm$ 0.00 &43.16 $\pm$ 0.00 &79.26 $\pm$ 0.00 &7848.86 $\pm$ 70.36 &4390.50 $\pm$ 263.65 
\\
& \multicolumn{1}{c|}{}
& MPLight &
141.80 $\pm$ 0.00 &36.38 $\pm$ 0.00 &520.81 $\pm$ 0.00 &7075.08 $\pm$ 23.24 &306.22 $\pm$ 25.09 
\\
& \multicolumn{1}{c|}{}
& AttendLight &
118.12 $\pm$ 2.78 &42.20 $\pm$ 0.86 &1313.50 $\pm$ 18.15 &4660.36 $\pm$ 88.32 &2669.69 $\pm$ 433.18 
\\
\cline{3-8}
& \multicolumn{1}{c|}{}
& UniLight &
96.65 $\pm$ 0.00 &27.18 $\pm$ 0.00 &22.03 $\pm$ 0.00 &2273.12 $\pm$ 246.13 &87.10 $\pm$ 24.52 
\\
\hline
\multirow{7}{*}{\rotatebox[origin=c]{90}{With}} & \multicolumn{1}{c|}{\multirow{7}{*}{\rotatebox[origin=c]{90}{UniComm}}}
& MA-A2C &
90.16 $\pm$ 1.55 &51.30 $\pm$ 0.96 &717.81 $\pm$ 42.64 &3972.36 $\pm$ 323.13 &170.02 $\pm$ 6.00 
\\
& \multicolumn{1}{c|}{} & MA-PPO &
89.48 $\pm$ 1.26 &51.49 $\pm$ 1.09 &733.61 $\pm$ 34.53 &3761.27 $\pm$ 197.65 &160.16 $\pm$ 4.23 
\\
& \multicolumn{1}{c|}{} & PressLight &
\textbf{77.83 $\pm$ 0.00} &32.86 $\pm$ 0.00 &1065.61 $\pm$ 0.00 &6173.58 $\pm$ 533.88 &424.60 $\pm$ 53.66 
\\
& \multicolumn{1}{c|}{}
& CoLight &
79.62 $\pm$ 0.00 &40.15 $\pm$ 0.00 &124.21 $\pm$ 0.00 &7596.31 $\pm$ 273.81 &1315.25 $\pm$ 649.06 
\\
& \multicolumn{1}{c|}{}
& MPLight &
96.48 $\pm$ 0.00 &29.62 $\pm$ 0.00 &26.73 $\pm$ 0.00 &5065.68 $\pm$ 226.48 &438.78 $\pm$ 125.50 
\\
& \multicolumn{1}{c|}{}
& AttendLight &
121.75 $\pm$ 3.65 &32.72 $\pm$ 1.02 &435.25 $\pm$ 39.52 &4787.83 $\pm$ 253.85 &2833.39 $\pm$ 775.06 
\\
\cline{3-8}
& \multicolumn{1}{c|}{}
& UniLight &
84.83 $\pm$ 0.00 &\textbf{25.53 $\pm$ 0.00} &\textbf{19.16 $\pm$ 0.00} &\textbf{73.46 $\pm$ 1.83} &\textbf{84.89 $\pm$ 1.32} 
\end{tabular}%

}
\caption{Evaluation result of average wait time (seconds).}
\label{tab:wait}
\end{table*}

\begin{table*}[htb]
\centering
\begin{tabular}{ccc||c|c|c|c|c}
\multicolumn{3}{c||}{Datasets} & {JN} & {HZ} & {NY} & {SH$_1$} & {SH$_2$} \\
\hline
\multirow{5}{*}{\rotatebox[origin=c]{90}{Traditional}} & \multicolumn{1}{c|}{\multirow{5}{*}{\rotatebox[origin=c]{90}{Algorithms}}} &
SOTL &
5369 $\pm$ 0 &2648 $\pm$ 0 &879 $\pm$ 0 &11459 $\pm$ 1193 &8769 $\pm$ 1042 
\\
& \multicolumn{1}{c|}{} & MaxPressure &
5330 $\pm$ 0 &2656 $\pm$ 0 &2642 $\pm$ 0 &6477 $\pm$ 216 &12245 $\pm$ 383 
\\
& \multicolumn{1}{c|}{} &
MaxBand &
5422 $\pm$ 6 &2574 $\pm$ 3 &1010 $\pm$ 21 &5316 $\pm$ 436 &4705 $\pm$ 99 
\\
& \multicolumn{1}{c|}{} &
TFP-TCC &
5607 $\pm$ 17 &2668 $\pm$ 5 &790 $\pm$ 32 &10267 $\pm$ 1458 &10341 $\pm$ 24 
\\
& \multicolumn{1}{c|}{} & MARLIN &
5028 $\pm$ 0 &2489 $\pm$ 0 &691 $\pm$ 0 &2895 $\pm$ 389 &3191 $\pm$ 909 
\\
\hline
\multirow{7}{*}{\rotatebox[origin=c]{90}{DRL}} & \multicolumn{1}{c|}{\multirow{7}{*}{\rotatebox[origin=c]{90}{Algorithms}}}
& MA-A2C &
5537 $\pm$ 15 &2708 $\pm$ 6 &1115 $\pm$ 41 &4344 $\pm$ 1701 &240 $\pm$ 27 
\\
& \multicolumn{1}{c|}{} & MA-PPO &
5059 $\pm$ 123 &2712 $\pm$ 4 &1168 $\pm$ 48 &4931 $\pm$ 2131 &5690 $\pm$ 743 
\\
& \multicolumn{1}{c|}{} & PressLight &
5610 $\pm$ 0 &2720 $\pm$ 0 &492 $\pm$ 0 &5066 $\pm$ 589 &14162 $\pm$ 175 
\\
& \multicolumn{1}{c|}{}
& CoLight &
5641 $\pm$ 0 &2714 $\pm$ 0 &425 $\pm$ 0 &1737 $\pm$ 154 &4169 $\pm$ 414 
\\
& \multicolumn{1}{c|}{}
& MPLight &
4653 $\pm$ 0 &2530 $\pm$ 0 &617 $\pm$ 0 &2114 $\pm$ 431 &4287 $\pm$ 71 
\\
& \multicolumn{1}{c|}{}
& AttendLight &
5380 $\pm$ 33 &2488 $\pm$ 22 &411 $\pm$ 26 &4470 $\pm$ 835 &3972 $\pm$ 174 
\\
\cline{3-8}
& \multicolumn{1}{c|}{}
& UniLight &
5626 $\pm$ 0 &2730 $\pm$ 0 &2686 $\pm$ 0 &8991 $\pm$ 695 &14746 $\pm$ 595 
\\
\hline
\multirow{7}{*}{\rotatebox[origin=c]{90}{With}} & \multicolumn{1}{c|}{\multirow{7}{*}{\rotatebox[origin=c]{90}{UniComm}}}
& MA-A2C &
5662 $\pm$ 12 &2711 $\pm$ 5 &1192 $\pm$ 50 &8091 $\pm$ 839 &14433 $\pm$ 46 
\\
& \multicolumn{1}{c|}{} & MA-PPO &
5662 $\pm$ 8 &2711 $\pm$ 4 &1249 $\pm$ 37 &7440 $\pm$ 691 &14453 $\pm$ 115 
\\
& \multicolumn{1}{c|}{} & PressLight &
5662 $\pm$ 0 &2726 $\pm$ 0 &561 $\pm$ 0 &2663 $\pm$ 381 &7323 $\pm$ 2052 
\\
& \multicolumn{1}{c|}{}
& CoLight &
\textbf{5675 $\pm$ 0} &2715 $\pm$ 0 &478 $\pm$ 0 &2124 $\pm$ 401 &5798 $\pm$ 413 
\\
& \multicolumn{1}{c|}{}
& MPLight &
5321 $\pm$ 0 &2725 $\pm$ 0 &721 $\pm$ 0 &1285 $\pm$ 486 &3854 $\pm$ 459 
\\
& \multicolumn{1}{c|}{}
& AttendLight &
5371 $\pm$ 10 &2730 $\pm$ 3 &671 $\pm$ 38 &6754 $\pm$ 657 &5254 $\pm$ 188 
\\
\cline{3-8}
& \multicolumn{1}{c|}{}
& UniLight &
5654 $\pm$ 0 &\textbf{2739 $\pm$ 0} &\textbf{2688 $\pm$ 0} &\textbf{16928 $\pm$ 3963} &\textbf{14538 $\pm$ 981} 
\end{tabular}%

\caption{Evaluation result of average throughput (number of vehicles).}
\label{tab:throughput}
\end{table*}

%\subsubsection{Overall Performance.}

We report the results corresponding to all four metrics below, including the average values and their standard deviations, in Tables~\ref{tab:travel} to~\ref{tab:throughput}. 
%
%Note that the performance of CoLight in three public datasets is different from the result 
%reported in the original paper where CoLight was introduced~\cite{wei2019colight}.
%Although we use the same evaluation metric as \cite{wei2019colight}, from the open-sourced codes, we find out that CoLight ignores the driving time when vehicles cross intersections and ignores the running time on the last outgoing road when measuring the travel time. 
%We re-include both time costs and hence the average travel time is slightly longer than that reported in \cite{wei2019colight}.
%
Note, the standard deviation of some experiments is zero. 
This is because in these datasets, the environment is determined, and DQN in testing is also determined. Consequently, the results remain unchanged in all 10 tests.

\subsubsection{Performance comparison without UniComm}

We first focus on the results without UniComm, i.e., all competitors communicate in their original way, and UniLight runs without any communication. The numbers in bold indicate the best performance. %We make several observations. 
%\qize{use footnote?} 
%
%
%they have some mistake in calculating the metric, which ignores the driving time when crossing intersections, as well as the running time on the last outgoing road, so the real average travel time is c larger.
We observe that RL based methods achieve better results than traditional methods in public datasets.
However, in a complicated environment like SH$_1$, agents may fail to learn a valid policy, and accordingly RL based methods might perform much worse than the traditional methods. 

UniLight performs the best 
%is able to achieve the best results consistently 
in almost all datasets, and it demonstrates significant advantages in complicated environments. 
%As compared with existing methods, UniLight, 
It improves the average performance by 8.2\% in three public datasets and 35.6\% in more complicated environments like SH$_1$/SH$_2$. 
%Note that in 
We believe the improvement brought by UniLight is mainly contributed by the following two features of UniLight. 
%there are two main reasons, as detailed below.
Firstly, UniLight divides the input state into traffic movements and uses the same model layer to generate corresponding hidden states for different traffic movements. As the layer is shared by all traffic movements regardless of traffic phases selected, the model parameters can be trained more efficiently. For example, the intersection shown in Figure 1 has 12 traffic movements. For every input state, the model layer is %equivalent to being 
actually trained 12 times.
Secondly, to predict Q-value of traffic phase $P$, we split hidden states of traffic movements into two groups based on their permissions and aggregate hidden states from same groups, so the model layer to predict Q-value can be used by different phases. This again implies that the layer is trained more times so it's able to learn better model weights. 

JN dataset is the only exception, where CoLight and PressLight perform slightly better than UniLight. This is because many roads in JN dataset share the same number of approaching vehicles, which makes the controlling easier, and all methods perform similarly. 
%Note that SOTL generates the best performance in SH$_1$, though UniLight performs much better than other competitors. However, once UniComm is enabled with the results to be reported next, UniLight will outperform SOTL in SH$_1$. 
%when UniComm is enabled. 
%However, in SH$_1$, UniLight fails to perform a better result compared with SOTL, which will be improved when introducing UniComm. 
%\qize{TODO when all data filled. The average improvement in public dataset is 50\%.}

%\subsubsection{The Impact of UniComm}
\subsubsection{The Impact of UniComm} 

As UniComm is universal for existing methods, we apply UniComm to the six representative RL based methods and re-evaluate their performance, with results listed in the bottom portion of Tables~\ref{tab:travel} to~\ref{tab:throughput}. 
We observe that UniLight again achieves the best performance consistently. In addition, almost all RL based methods (including UniLight) are able to achieve a better performance with UniComm.
%, and their average improvement ranges from \qize{xxx\% to xxx\%}. 
%Note that for MPLight in SH$_2$, the result with UniComm is not good as original MPLight, which decreases the average improvement. The reason is its performance in complicated envirionment is not stable (e.g. the variance is 964.2). 
This is because UniComm predicts the approaching vehicle number on $R_{i,j}$ mainly by the hidden states of traffic movements covering $R_{i,j}$. Consequently, $A_i$ is able to produce predictions for neighboring $A_j$ based on more important/relevant information.
As a result, neighbors will receive customized results, which allow them to utilize the information in a more effective manner.  
In addition, agents only share predictions with their neighbors so the communication information and the observation dimension remain small. This allows existing RL based methods to outperform their original versions whose communications are mainly based on hidden states. 
%We also want to highlight that 
Some original methods perform worse with UniComm in certain experiments, e.g. PressLight in SH$_1$. This is because these methods have unstable performance on some datasets with very large variance, which make the results unstable. Despite of a small number of outliers, UniComm makes consistent boost on all existing methods.
% For example, the variance of MPLight in SH$_2$ is as high as \qize{???}.
%We also want to highlight that original MPLight outperforms its version with UniComm under SH$_2$, the only exceptional case. 
%This is because under complicated environments, the performance of MPLight is very unstable and the variance is as high as 964.2.%, ii) under HZ, the  performances difference is too small.
%and make them easier to use the information.
%And as we only send the prediction, the communication information remains small, and the observation dimension is not largely increased. So the convergence speed is quicker than passing hidden states. 

\subsubsection{Phase Prediction Target Evaluation}
As mentioned previously, instead of directly using current phase action to calculate phase prediction loss $L_p$, we use actions stored in replay buffer. 
To evaluate its effectiveness, we plot the 
%phase prediction loss 
curve of $L_p$ in Figure~\ref{fig:phaselossall}. %Due to space limitation, we only show the curve of datasets JN and HZ.%, and report the curves of remaining datasets in the supplementary material. 
\emph{Current} and \emph{Replay} refer to the use of the action taken by the current network and that stored in the replay buffer respectively when calculating $L_p$.
%As we run four independent experiments for every combination of methods and datasets, 
The curve represents the volume prediction loss, i.e. the prediction accuracy during the training process. %\baihua{Previously, we mentioned that we run exp four times and then picked the best one. We then run the best one 10 times to report the avg performance. Contradicts with what we present here. }\qize{For experiment results, we use the best model to evaluate; and here is the training curve, so we give average of them. To make experiment settings consistent, I will re-draw this figure with the curve of best model, and put average loss curve figure into appendix.}
%\baihua{Which four? Needs explanation.} 
%From the curve, 
We can observe that when using stored actions as the target, the loss becomes smaller, i.e., it has learned better phase predictions. 
The convergence speed of phase prediction loss with stored actions is slower at the beginning of the training. This is because to perform more exploration, most actions are random at the beginning, which is hard to predict. % make the prediction very difficult. 
%which is hard to predict. 

%\subsubsection{UniComm vs. Hidden State}
\subsubsection{UniComm vs. Hidden States}

To further evaluate the effectiveness of UniComm from a different angle, we introduce another version of UniLight that uses a 32-dimension hidden state for communication, the same as \cite{wei2019colight}. 
In total, there are three different versions of UniLight evaluated, i.e., \emph{No Com.}, \emph{Hidden State}, and \emph{UniComm}.
%
%and the results are listed in Table \ref{tab:unicommvshs}. 
As the names suggest, \emph{No Com.} refers to the version without sharing any information as there is no communication between agents; \emph{Hidden State} refers to the version sharing 32-dimension hidden state; and \emph{UniComm} refers to the version that implements UniComm. 
The average travel time of all the vehicles under these three variants of UniLight is listed in Table \ref{tab:unicommvshs}.
Note, \emph{Hidden State} shares the most amount of information and is able to improve the performance, as compared with version of \emph{No Com.}. 
Unfortunately, the amount of information shared is \emph{not} proportional to the performance improvement it can achieve. 
For example, UniLight with UniComm performs better than the version with Hidden State, although it shares less amount of information. 
This further verifies our argument that the content and the importance of the shared information is much more important than the amount. 

%
%As more information has been transmitted, the performance should at least remains same considering the total amount of information. However, although the performance is better than UniLight without any communication, in all datasets the performance is worse than UniLight with UniComm, which proves UniComm is better than heuristic communication.

\section{Conclusion}
In this paper, we propose a novel communication form UniComm for decentralized multi-agent learning based M-TSC problem.  
It enables each agent to share the prediction of approaching vehicles with its neighbors via communication.
%Supported by traffic flow theories, UniComm can well represent the impacts from other agents.
We also design UniLight to predict Q-value based on UniComm.
Experimental results demonstrate that UniComm is universal for existing M-TSC methods, and UniLight outperforms existing methods in both simple and complex environments. 

\bibliographystyle{named}
\bibliography{ijcai22-short}

\end{document}